\documentclass[11pt]{article}
\usepackage[final]{acl}

\usepackage{times}
\usepackage{latexsym}
\usepackage{fontawesome}

\usepackage[T1]{fontenc}

\usepackage[utf8]{inputenc}

\usepackage{microtype}

\usepackage{inconsolata}

\usepackage{graphicx}
\usepackage{booktabs}
\usepackage{siunitx}
\usepackage[table]{xcolor}  
\usepackage{caption}
\usepackage{multirow}
\usepackage{amsfonts}
\usepackage{amsmath}
\usepackage{amssymb}
\usepackage{pifont}
\usepackage{algorithm}
\usepackage{algorithmic}
\usepackage{tcolorbox}

\newcommand{\RMiPO}{\texttt{RMiPO}}

\title{Intrinsic Mutual Information as a Modulator for Preference Optimization}

\author{
Peng Liao$^{1}$ \quad Peijia Zheng$^{1}$ \quad Lingbo Li$^{2}$ \quad
Shangsong Liang$^{1}$\footnotemark[1] \quad
Lin Chen$^{3}$\thanks{Corresponding authors.}\\[3pt]
$^{1}$Sun Yat-sen University \quad $^{2}$University of Warwick \quad $^{3}$Macao Polytechnic University \\[3pt]
\texttt{liavonpenn@gmail.com \quad liangshangsong@gmail.com \quad lchen@mpu.edu.mo}
}

\begin{document}
\maketitle
\begin{abstract}
Offline preference optimization methods, such as Direct Preference Optimization (DPO), offer significant advantages in aligning Large Language Models (LLMs) with human values. 
However, achieving optimal performance with these methods typically involves additional hyperparameter tuning, resulting in substantial time overhead.
Although prior work has proposed a range of improvements, these methods remain limited in effectiveness and have not fully eliminated reliance on hyperparameter tuning.
In this work, we propose \RMiPO, a lightweight and efficient framework for offline preference optimization. \RMiPO~leverages intrinsic \underline{\textbf{R}}esponse-level \underline{\textbf{M}}utual \underline{\textbf{i}}nformation for \underline{\textbf{P}}reference \underline{\textbf{O}}ptimization with hyperparameter modulation, dynamically decoupling preference contributions at negligible additional computational cost. Extensive experimental results demonstrate that \RMiPO~achieves consistently superior performance over existing methods while reducing training overhead by more than 15\%. Our code is available at \url{https://github.com/liavonpenn/rmipo}.
\end{abstract}

\section{Introduction}

Large Language Models (LLMs) demonstrate remarkable generation and comprehension capabilities across diverse NLP tasks~\cite{llama2025, deepseek2025, gpt2025}. However, when faced with complex tasks such as mathematical analysis and logical reasoning, LLMs may take incorrect reasoning paths or generate inconsistent conclusions. To address this issue, researchers have proposed Reinforcement Learning from Human Feedback (RLHF)~\cite{ouyang2022training}. This approach fine-tunes pretrained LLMs using carefully curated human preference data, improving their reasoning fidelity and alignment with human values.

Although RLHF substantially enhances LLM performance, its intricate reward modeling pipeline and significant computational demands hinder scalability. To mitigate these limitations while preserving alignment quality, researchers have introduced offline preference optimization methods such as DPO~\cite{rafailov2023direct}, IPO~\cite{azar2024general}, and KTO~\cite{ethayarajh2024kto}. These approaches optimize preferences directly through the likelihood outputs of the policy model, thereby obviating the need for an explicit reward model. Building upon this paradigm, SimPO~\cite{meng2024simpo} further eliminates dependence on a reference model, yielding an approximately 20\% reduction in training time for single-round preference optimization relative to DPO. Empirically, these methods perform comparably to, and in some cases surpass, RLHF. Nevertheless, as noted in \citet{huggingface2024_ft},
fully realizing their potential still necessitates careful tuning of several key hyperparameters. This process is nontrivial, as each alignment run is costly and configuration changes often require re-establishing optimal hyperparameters.

\begin{figure}[t]
\centering
\includegraphics[width=\columnwidth]{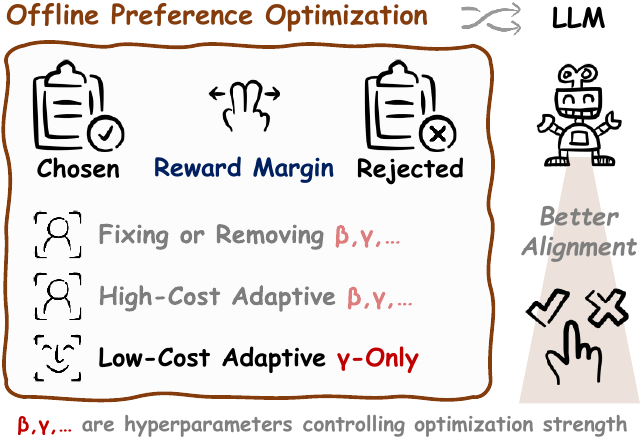} 
\caption{Unlike existing static or high-cost adaptive methods, \RMiPO~leverages intrinsic mutual information for modulation, enabling adaptive $\gamma$-only control to balance training efficiency and alignment performance.}
\label{fig-1}
\vspace{-1.5em}
\end{figure}

Recent solutions can be broadly divided into two types: one embeds dynamic mechanisms in preference optimization~\cite{wu2024beta, lee2025kl, wu2025alphadpo} to adaptively adjust key hyperparameters, while the other fundamentally alters the optimization paradigm, as exemplified by~\citet{xiao2025simper}, thereby removing hyperparameter dependence. Both approaches exhibit distinct theoretical and empirical advantages. However, the former often requires additional monitoring modules or complex control logic, adding extra training overhead, whereas the latter may degrade under preference drift or inconsistent data~\cite{yang2024not,gorbatovski2024learn,zeng2024token}. This raises a critical research question in LLM alignment: \emph{Is it feasible to develop an efficient method for offline preference optimization?}

In this paper, we provide an affirmative answer to this question. As illustrated in Figure~\ref{fig-1}, we propose \RMiPO, a lightweight and efficient framework for offline preference optimization. At its core is a dynamic modulation mechanism derived from intrinsic mutual information, enabling \RMiPO~to achieve instance-level adaptive tuning of hyperparameters. 
Through systematic theoretical analysis, we find that the hyperparameter $\gamma$, which controls the reward margin, plays a dominant role in preference optimization, while the intrinsic mutual information of responses provides a principled basis for its adaptive adjustment. 
Building on the above insight, \RMiPO~adaptively adjusts $\gamma$ during optimization, enabling decoupled modeling of preference contributions at negligible computational overhead.

The primary contributions of this work are as follows: 
(1) \textbf{Lightweight Framework}. We propose \RMiPO, a novel lightweight offline preference optimization framework that leverages intrinsic mutual information to enable instance-level adaptive hyperparameter control.
(2) \textbf{Insightful Analysis}. We conduct a comprehensive examination of the roles of hyperparameters in offline preference optimization framework and offer meaningful insights that can guide future practice and research.
(3) \textbf{Superior Performance}. Through extensive evaluations on benchmarks such as AlpacaEval 2 and MT-Bench, \RMiPO~reduces training costs by approximately 15–20\% while achieving performance clearly superior to state-of-the-art baselines.

\section{Preliminary}

\textbf{Problem Formulation}. This paper focuses on the offline preference alignment task. We assume access to a dataset $\mathbb{D} = \{(x^{(i)}, y^{(i)}_w, y^{(i)}_l)\}_{i=1}^N$ of $N$ preference triplets, where each sample consists of a prompt $x^{(i)}$ and a pair of responses $y^{(i)}_w$ (preferred) and $y^{(i)}_l$ (less-preferred). 
Each preference triplet is labeled by a preference oracle, such as humans or LLM-based evaluators, which identifies the preferred response.
To learn from $\mathbb{D}$, \citet{christiano2017deep}~adopted the Bradley–Terry (BT) model~\cite{bradley1952rank} as the basis for learning. In the BT formulation, the probability that $y_w$ is preferred to $y_l$ given $x$ is:
\begin{equation}
\mathbb{P}(y_w \succ y_l \mid x)= \sigma(r(x,y_w)-r(x,y_l)),
\end{equation}
where $\sigma(\cdot)$ is the logistic function and $r(x,y)$ is the implicit quality score induced by the current policy. Under this setup, we directly learn a parameterized policy model $\pi_\theta(y \mid x)$, using the preference signal implicitly encoded in $r(x,y)$ to guide updates. The policy is optimized solely on $\mathbb{D}$ and progressively aligns its generation with the oracle’s preferences.

\noindent \textbf{Direct Preference Optimization} (DPO)~\cite{rafailov2023direct} is a widely used method for offline preference optimization. Its core idea is to leverage the log-likelihood of $\pi_{\theta}$ to construct an implicit, preference-aligned reward by relating the optimal policy in closed form to a reference policy $\pi_{\mathrm{ref}}$. Specifically, DPO defines the reward as
$
r(x, y) = \beta \log \frac{\pi_{\theta}(y \mid x)}{\pi_{\mathrm{ref}}(y \mid x)} + \beta \log Z(x),
$
where $Z(x)$ is a partition function independent of $y$. 
The hyperparameter $\beta > 0$ scales the log-likelihood difference to control KL regularization and keep $\pi_{\theta}$ close to $\pi_{\mathrm{ref}}$. Under this reward, the objective becomes
\begin{equation}
\begin{aligned}
&\mathcal{L}_{\mathrm{DPO}}
    = -\mathbb{E}_{(x, y_w, y_l) \sim \mathbb{D}}
       \left[ \log \sigma( \Delta r ) \right], \text{where}\\
&\Delta r
    = \beta\log \frac{\pi_{\theta}(y_w | x)}{\pi_{\mathrm{ref}}(y_w | x)}
     - \beta\log \frac{\pi_{\theta}(y_l | x)}{\pi_{\mathrm{ref}}(y_l | x)} .
\label{eq-dpo}
\end{aligned}
\end{equation}
This objective encourages increasing the preferred response’s log-likelihood ratio and decreasing the less-preferred one relative to the reference policy.

\noindent \textbf{Simple Preference Optimization} (SimPO)~\cite{meng2024simpo} removes DPO’s reliance on a reference model and mitigates length bias in autoregressive generation.
It introduces a length-normalized reward $\tilde{r}(x, y) = \frac{\beta}{|y|} \log \pi_\theta(y | x)$, where $|y|$ is the response length. 
To further amplify preference signal, SimPO incorporates a margin hyperparameter $\gamma > 0$ within the BT formulation, requiring the preferred response to exceed the less-preferred one by at least $\gamma$. This yields the following objective:
\begin{equation}
\mathcal{L}_{\mathrm{SimPO}} = -\mathbb{E}_{(x, y_w, y_l) \sim \mathbb{D}} \left[ \log \sigma \left( \Delta \tilde{r}  - \gamma \right) \right],
\label{eq-simpo}
\end{equation}
where $\Delta \tilde{r} = \tilde{r}(x, y_w) - \tilde{r}(x, y_l)$ by definition.

Alignment performance under the above and other offline preference objectives (Appendix~\ref{sec:methods}) is highly sensitive to hyperparameters, especially $\beta$ and $\gamma$, whose optimal values demand costly tuning. 
To this end, we introduce \RMiPO, with an instance-level adaptive mechanism that significantly mitigates hyperparameter tuning overhead.


\section{Methodology}

In this section, we address two core design questions: (1) Which hyperparameters necessitate adaptive adjustment, and (2) how to design effective mechanisms for such adaptation. $\S$\ref{set3.1} presents two key insights, and in $\S$\ref{set3.2} we introduce \RMiPO, a novel offline preference optimization framework that effectively resolves both of these challenges.

\subsection{Motivation and Key Insights}\label{set3.1}

The performance of offline preference optimization depends on hyperparameter choice, particularly the trade-off between policy regularization and reward scaling. Through extensive experiments and theoretical analysis, we have distilled two key insights.

\begin{figure}[t]
\centering
\includegraphics[width=\columnwidth]{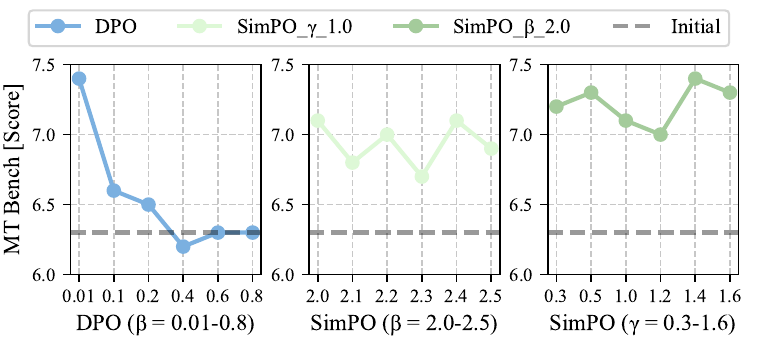} 
\caption{Hyperparameter sensitivity of Mistral-7B-Base on MT-Bench. DPO is evaluated across varying $\beta$, while SimPO is tested under two settings: varying $\beta$ with fixed $\gamma=1.0$, and varying $\gamma$ with fixed $\beta=2.0$. The dashed line denotes the SFT-initialized baseline.}
\label{fig-mtbench}
\vspace{-1em}
\end{figure}

\noindent \emph{\textbf{Insight I}. In offline preference optimization, optimal performance is primarily determined by $\gamma$.} 
As shown in Appendix Table~\ref{tab:baseline}, the offline preference learning paradigm involves more than two hyperparameters, but its core remains $\beta$, which scales reward differences, and $\gamma$, which controls reward margins. We fine-tuned Mistral-7B-Base and conducted sensitivity analyses on MT-Bench~\cite{zheng2023judging}. Although training curves and in-domain validation are similar across configurations, MT-Bench scores vary sharply (Figure~\ref{fig-mtbench}). In DPO, except for $\beta = 0.01$, which performs relatively well but is not a universal solution~\cite{wu2025alphadpo}, all other $\beta$ settings lead to substantial performance degradation, in some cases falling even below the SFT initialization baseline. In contrast, in SimPO, adjusting either $\beta$ or $\gamma$ improves performance, with $\gamma$ exerting an especially pronounced positive effect.

While incorporating $\gamma$ into DPO yields negligible improvements \cite{meng2024simpo}, it is indispensable for SimPO, as its removal degrades performance.
This motivates us to reexamine the DPO objective (Eq.~\ref{eq-dpo}), we observe that although it lacks $\gamma$ explicitly, the reference model $\pi_{\mathrm{ref}}$ induces an \textbf{instance-level implicit margin}. Specifically, the DPO loss can be rewritten analogously to SimPO:
\begin{align}
&\mathcal{L}_{\mathrm{DPO}} = 
\log \sigma \Big(
    \beta \log \pi_\theta(y_w|x)
    - \beta \log \pi_\theta(y_l|x) \nonumber \\
&    - \underbrace{\big(\beta \log \pi_{\mathrm{ref}}(y_w|x) - \beta \log \pi_{\mathrm{ref}}(y_l|x)\big)}_{\text{Instance-level~Implicit~Margin}~\gamma_{\mathrm{ref}}}
\Big). \label{eq-dpo_v1}
\end{align}
Based on this observation, we unify the optimization objectives of DPO and SimPO as $\mathcal{L}=-\mathbb{E}\left[ \log \sigma(\Delta \mathbb{R} - \gamma)\right]$, where $\Delta \mathbb{R} = \beta \Delta \log$ denotes the reward difference induced by the log-likelihood gap of the policy model. This formulation encompasses $\Delta r$ in DPO and $\Delta \tilde{r}$ in SimPO.

Under this unified framework, policy updates follow the gradient descent rule $\theta \leftarrow \theta - \eta \, \nabla_{\theta} \mathcal{L}(\theta; \mathbb{D})$, where $\eta > 0$ denotes the learning rate. The gradient of the objective can be written explicitly as
\begin{equation}
\nabla_{\theta} \mathcal{L} = -\beta \, \mathbb{E} \left[ \big(1 - \sigma(\Delta \mathbb{R} - \gamma)\big)\nabla_{\theta} \Delta \mathbb{R} \right].
\end{equation}
It is noteworthy that $\beta$ serves merely as a scalar multiplier rescaling gradients, whereas $\gamma$ appears directly inside the sigmoid and directly determines each sample’s gradient-sensitivity region. Under the first-order optimality condition, stationary points satisfy $\nabla_{\theta} \mathcal{L} = 0$, which, after substituting the gradient and noting $\beta>0$, is equivalent to $\mathbb{E} \left[ \big(1 - \sigma(\beta\Delta \log - \gamma)\big)\nabla_{\theta} \Delta \log \right] = 0$. Thus, the stationary set is essentially governed by $\gamma$, as it modulates per-sample weighting via $\sigma(\beta\Delta\log-\gamma)$, whereas $\beta$ only changes the sigmoid’s steepness without shifting stationary points. These observations clearly indicate that $\gamma$ should serve as the \textbf{primary parameter for adaptive tuning} in offline preference optimization framework.

\begin{figure}[t]
\centering
\includegraphics[width=\columnwidth]{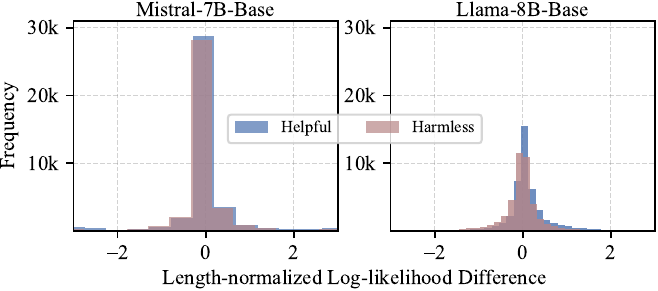} 
\caption{Near-zero, overlapping log-likelihood differences provide a weak signal for adaptive tuning.}
\label{fig-analyes}
\vspace{-1em}
\end{figure}

\noindent \emph{\textbf{Insight II}. Modulating $\gamma$ via mutual information differences offers a reliable, lightweight scheme.}
An intuitive and previously explored strategy~\cite{wu2024beta} is to adaptively adjust $\gamma$ using log-likelihood differences computed by the policy model over preference pairs. However, our empirical analysis of the Anthropic-HH dataset~\cite{bai2022training}, as shown in Figure~\ref{fig-analyes}, reveals several issues. For both Mistral-7B-Base and Llama-8B-Base, the log-likelihood differences in "Helpful" and "Harmless" categories cluster tightly near zero, exhibit asymmetry, and substantially overlap between preferred and non-preferred samples. Moreover, their distributional shapes vary markedly across models and preference. 
These properties indicate that likelihood differences are a weak, sensitive, low–signal-to-noise signal with limited effectiveness for adaptive mechanisms built upon them.

This phenomenon motivates a deeper examination of intrinsic limitations in log-likelihood differences. 
Although $\log \pi_{\theta}(y \mid x)$ is intended to measure semantic relevance, it in practice reflects a composite signal entangled with multiple confounding factors.
Applying Bayes' rule and taking the logarithm yields: $\log\pi_{\theta }(y|x)=\log\frac{\pi_{\theta }(x|y)\pi_{\theta }(y)}{\pi_{\theta }(x)}$. Here, $\log \pi(x|y)$ captures the semantic coupling between the response and the prompt (the "reverse relevance" to be optimized), while $\log \pi(y)$ reflects the response's prior popularity. The term $\log \pi(x)$ is a constant that does not affect preference ranking. If the adjustment mechanism is based on the policy model’s log-likelihood gap $\Delta \log=\log\frac{\pi_\theta(y_w|x)}{\pi_\theta(y_l|x)}$:
\begin{equation}
\gamma \propto \Delta \log  = \underbrace{\log \frac{\pi_\theta (x|y_w)}{\pi_\theta(x|y_l)}}_{\text{Relevance Gap}} + \underbrace{\log \frac{\pi_\theta(y_w)}{\pi_\theta(y_l)}}_{\text{Popularity Bias}},
\end{equation}
then the resulting signal inevitably entangles true relevance differences with popularity bias. When $y_w$ is a generic, fluent but low-informative ``boilerplate'' response such as ``That's a great question''~\cite{li2016diversity}, $\log \pi(y_w)$ becomes large. Consequently, the policy model tends to favor such safe yet uninformative responses. 
Although sequence length replaces reference-model constraints~\cite{wu2024beta, xiao2025simper}, it remains insufficient to mitigate popularity bias.

In contrast, when incorporating a reference model, the adjustment signal is formulated as $\Delta \log=\log\frac{\pi_\theta(y_w|x)}{\pi_\mathrm{ref}(y_w|x)}- \log\frac{\pi_\theta(y_l|x)}{\pi_\mathrm{ref}(y_l|x)}$. When the policy and reference models have similar language modeling capabilities (reference is typically initialized policy), this formulation naturally cancels prior popularity bias. Specifically, under the assumption that $\pi_\theta \approx \pi_{\mathrm{ref}}$, we have $\log \pi_\theta(y) - \log \pi_{\mathrm{ref}}(y) \approx 0$. This observation not only explains DPO’s robustness to generic responses but also reveals the reason behind the effectiveness of recent adaptive tuning methods~\cite{lee2025kl, wu2025alphadpo}. However, this comes at the cost of additional model storage and inference overhead.

Existing strategies face an \textbf{inherent trade-off}: either they fail to fully eliminate response-prior contamination in the adjustment signal, or rely on reference models, which impose additional computational and storage overhead. This raises a critical question: is there an adjustment signal capable of decoupling semantic relevance from response priors without relying on a reference model? We propose to leverage pointwise mutual information (PMI), a well-established information-theoretic measure \cite{chen2024language, liu2025pointwise, xiao2025infopo}, as the adaptive signal for margin control. Formally, under the policy model $\pi_\theta$, PMI is defined as $\mathrm{PMI}(y| x) = \log \frac{\pi_\theta(y | x)}{\pi_\theta(y)}$. The resulting adjustment signal is $\Delta \mathrm{pmi}=\mathrm{PMI} (y_w|x)-\mathrm{PMI}(y_l|x)$. Substituting the Bayes decomposition yields:
\begin{equation}\label{eq:pmi}
\gamma \propto \Delta \mathrm{pmi}=\log \pi_\theta (x|y_w) - \log\pi_\theta (x|y_l).
\end{equation}
This analysis provides an information-theoretic justification for using mutual-information differences as the adjustment signal for the hyperparameter $\gamma$; the full derivation is provided in Appendix~\ref{sec:proofs}.

\subsection{Proposed Framework: \RMiPO}\label{set3.2}


These insights establish two principles: treating $\gamma$ as the key adaptive variable and updating it via mutual information differences. To this end, we propose \RMiPO, a novel optimization framework.

\begin{figure}[t]
\centering
\includegraphics[width=\columnwidth]{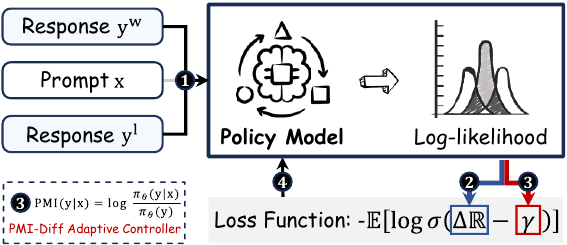} 
\caption{\RMiPO~training workflow: four-step reference-free policy optimization using mutual information.}
\label{fig-method}
\vspace{-1em}
\end{figure}

\noindent \textbf{Overview}. Figure~\ref{fig-method} illustrates the \RMiPO~training workflow. Given a prompt $x$ and a preference pair $(y_w, y_l)$, the policy model $\pi_\theta$ computes pointwise mutual information for each response, denoted as $\mathrm{PMI}(y|x).$ Unlike the raw log-likelihood difference, which entangles semantic relevance with response priors, the PMI difference $\Delta \mathrm{pmi}$ more directly disentangles conditional relevance from marginal popularity. This signal (Step~\ding{184}) thus enables reliable and efficient adaptation of the reward margin $\gamma$ and offers advantages over prior work. Details are provided in the following sections.

\noindent \textbf{Objective}. In offline preference optimization, the policy $\pi_\theta$ increases $\Delta \mathbb{R}$ by maximizing the log-likelihood ratio between preferred and less-preferred responses, defining the gradient direction for preference learning. The reward margin $\gamma$ gates updates based on semantic separability: samples with $\Delta \mathbb{R} > \gamma$ are suppressed, others contribute. To this end, \RMiPO~employs the pointwise mutual information difference, $\Delta \mathrm{pmi}(x,y_w,y_l) = \mathrm{PMI}(y_w|x) - \mathrm{PMI}(y_l|x)$, as an intrinsic metric for semantic separability. This signal eliminates interference from non-semantic factors, such as prior frequencies. Building on this, \RMiPO~introduces a key design: the reward margin $\gamma$ is negatively correlated with $\Delta \mathrm{pmi}$. Under this mechanism, semantically distinct pairs (high $\Delta \mathrm{pmi}$) receive smaller margins to promote saturation and reduce redundancy. Conversely, ambiguous samples (low $\Delta \mathrm{pmi}$) retain larger margins to sustain gradients on semantic differences at the decision boundary.

As shown in Fig.~\ref{fig-analyes}, in early training the policy $\pi_\theta$ lacks reliable preference discrimination, resulting in an effectively ungated margin regime ($\Delta\mathbb{R}<\gamma$) where most samples remain gradient-active. As learning progresses, the margin $\gamma$ progressively reallocates gradients, yielding a smooth transition from coarse discrimination to fine-grained alignment. To model this nonlinear dependency, we define the dynamic margin as a rectified exponential decay of the pointwise mutual information difference:
$\gamma(x, y_w, y_l) = \gamma_{\min} + (\gamma_{\max} - \gamma_{\min}) \cdot \exp(-\max(0, \Delta \mathrm{pmi}))$.
The rectification introduces asymmetry: when $\Delta\mathrm{pmi}\le0$, the margin is fixed at $\gamma_{\max}$, enforcing a strong constraint that continuously penalizes inverted semantic preferences; otherwise, the margin decays exponentially with increasing confidence. This design preserves optimization signals in low- and medium-confidence regions while preventing premature margin collapse.

Moreover, inspired by SimPO~\cite{meng2024simpo}, we incorporate length normalization into the reward signal to further stabilize training dynamics. Notably, as shown in the Appendix Table~\ref{tab:LN-AE2}, even in settings without length normalization, our method remains effective and consistently improves performance. Integrating the above design, the final loss function of \RMiPO, $\mathcal{L}_{\text{\RMiPO}}$, is formulated as:
\begin{equation}\label{eq:loss}
-\mathbb{E}_{(x, y_w, y_l) \sim \mathbb{D}} \left[ 
\log \sigma \big( \Delta \mathbb{R} - \gamma(x, y_w, y_l) \right) 
\big].
\end{equation}
Here, $\Delta \mathbb{R} = \frac{\beta}{|y_w|} \log \pi_\theta(y_w | x) - \frac{\beta}{|y_l|} \log \pi_\theta(y_l | x)$ represents the preference score difference after length normalization. By explicitly coupling $\Delta \mathrm{pmi}$ with $\gamma$, \RMiPO~achieves adaptive perception and response to the preference structure. This framework not only has a solid theoretical foundation but also demonstrates computational feasibility, offering a new paradigm for offline preference optimization that combines efficiency and robustness. Further details and pseudocode are in the Appendix~\ref{sec:details}.


\setlength{\tabcolsep}{2pt}
\begin{table*}[!t]
\centering
\small
\caption{AlpacaEval 2 and MT-Bench results under four settings. LC and WR denote the widely adopted length-controlled and raw win rates, respectively. \RMiPO~consistently achieves measurable gains across various settings.}
\label{tab:ae2_mtb}
\vskip -1em
\resizebox{\textwidth}{!}{
\begin{tabular}{lcccccccccccc}
\toprule
\multirow{3}{*}{\textbf{Method}} 
& \multicolumn{3}{c}{\textbf{Mistral-7B-Base}} 
& \multicolumn{3}{c}{\textbf{Mistral-7B-Instruct}} 
& \multicolumn{3}{c}{\textbf{Llama3-8B-Base}} 
& \multicolumn{3}{c}{\textbf{Llama3-8B-Instruct}} 
\\
\cmidrule(lr){2-4}\cmidrule(lr){5-7}\cmidrule(lr){8-10}\cmidrule(lr){11-13}
& \multicolumn{2}{c}{\textbf{AlpacaEval 2}} 
& \textbf{MT-Bench} 
& \multicolumn{2}{c}{\textbf{AlpacaEval 2}} 
& \textbf{MT-Bench} 
& \multicolumn{2}{c}{\textbf{AlpacaEval 2}} 
& \textbf{MT-Bench} 
& \multicolumn{2}{c}{\textbf{AlpacaEval 2}} 
& \textbf{MT-Bench} \\
\cmidrule(lr){2-3}\cmidrule(lr){4-4}
\cmidrule(lr){5-6}\cmidrule(lr){7-7}
\cmidrule(lr){8-9}\cmidrule(lr){10-10}
\cmidrule(lr){11-12}\cmidrule(lr){13-13}
& {\scriptsize \bf LC (\%)} 
& {\scriptsize \bf WR (\%)} 
& {\scriptsize \bf GPT-4} 
& {\scriptsize \bf LC (\%)}  
& {\scriptsize \bf WR (\%)} 
& {\scriptsize \bf GPT-4} 
& {\scriptsize \bf LC (\%)}  
& {\scriptsize \bf WR (\%)} 
& {\scriptsize \bf GPT-4} 
& {\scriptsize \bf LC (\%)}  
& {\scriptsize \bf WR (\%)} 
& {\scriptsize \bf GPT-4} \\
\midrule
Initial 
& 8.4 & 6.2 & 6.3 
& 17.1 & 14.7 & 7.5 
& 6.2 & 4.6 & 6.6 
& 26.0 & 25.3 & 8.1
\\
\midrule
DPO  
& 15.1 & 12.5 & 7.3 
& 26.8 & 24.9 & 7.6 
& 18.2 & 15.5 & 7.7 
& 40.3 & 37.9 & 8.0
\\
SLiC 
& 10.9 & 8.9 & 7.4 
& 24.1 & 24.6 & 7.8
& 12.3 & 13.7 & 7.6 
& 26.9 & 27.5 & 8.1
\\
IPO  
& 11.8 & 9.4 & 7.2 
& 20.3 & 20.3 & 7.8 
& 14.4 & 14.2 & 7.4 
& 35.6 & 35.6 & \textbf{8.3}
\\
KTO  
& 13.1 & 9.1 & 7.0 
& 24.5 & 23.6 & 7.7 
& 14.2 & 12.4 & \textbf{7.8} 
& 33.1 & 31.8 & 8.2
\\
CPO 
& 9.8 & 8.9 & 6.8 
& 23.8 & 28.8 & 7.5 
& 10.8 & 8.1 & 7.4 
& 28.9 & 32.2 & 8.0
\\
SimPO 
& 21.5 & 20.8 & 7.3 
& 32.1 & 34.8 & 7.6 
& 22.0 & 20.3 & 7.7 
& 44.7 & 40.5 & 8.0
\\
\midrule
$\alpha$-DPO
& 21.0 & 20.5 & \textbf{7.5} 
& 32.3 & 32.6 & \textbf{8.0}
& 24.6 & 22.7 & 7.7 
& 46.6 & 38.1 & 8.1
\\
$\beta$-DPO
& 15.7 & 16.8 & 7.3 
& 23.6 & 20.4 & 7.7
& 23.5 & 20.7 & 7.7 
& 43.4 & 38.2 & 8.0
\\
$\epsilon$-DPO
& 22.0 & 20.8 & 7.4 
& 35.6 & 29.6 & 7.8
& \textbf{26.0} & 23.4 & 7.6 
& 46.4 & 44.9 & 8.0
\\
SimPER 
& \textbf{22.4} & 21.3 & \textbf{7.5} 
& 37.8 & \textbf{39.5} & 7.8 
& 25.2 & 22.9 & 7.7 
& 48.5 & 45.7 & 8.2
\\
\midrule
\RMiPO 
& 22.3 & \textbf{21.4} & \textbf{7.5} 
& \textbf{38.1} & 37.0 & 7.8
& 25.2 & \textbf{23.6} & \textbf{7.8}
& \textbf{49.1} & \textbf{46.3} & \textbf{8.3}
\\
\bottomrule
\end{tabular}
}
\vspace{-0.5em}
\end{table*}


\section{Experiments}

\subsection{Experimental Setups}

Following~\cite{xiao2025simper}, we perform preference optimization in two settings (Base and Instruct) across two model families, namely Llama3-8B~\cite{dubey2024llama} and Mistral-7B~\cite{jiang2023mistral7b}. In the Base setting, the initial policy is constructed using the supervised fine-tuning (SFT) pipeline of Zephyr~\cite{tunstall2023zephyr}. In the Instruct setting, we directly adopt the officially released instruction-tuned models as the initial policy, skipping the SFT stage. Subsequently, these models are further optimized for preferences on the UltraFeedback dataset~\cite{cui2023ultrafeedback}. All experiments are implemented using the alignment-handbook~\cite{Tunstall_The_Alignment_Handbook} and conducted on four NVIDIA A800 (80GB) GPUs with batch size 128.

The resulting policy models are evaluated on multiple authoritative benchmarks, including the Open LLM Leaderboard v2~\cite{open-llm-leaderboard-v2, eval-harness}, AlpacaEval2~\cite{alpaca_eval}, and MT-Bench~\cite{zheng2023judging}. To further assess safety alignment, we conduct targeted fine-tuning and evaluation on the Anthropic-HH (Helpful and Harmless) dataset~\cite{bai2022training}. 
\RMiPO~is compared with \textbf{the standard offline approaches} (DPO~\cite{rafailov2023direct}, SLiC~\cite{zhao2023slic}, IPO~\cite{azar2024general}, KTO~\cite{ethayarajh2024kto}, CPO~\cite{xu2024contrastive}, SimPO~\cite{meng2024simpo}), and \textbf{the solutions addressing hyperparameter sensitivity} ($\alpha$-DPO~\cite{wu2025alphadpo}, $\beta$-DPO~\cite{wu2024beta}, $\epsilon$-DPO~\cite{lee2025kl}, SimPER~\cite{xiao2025simper}).
For each baseline, we conduct extensive tuning within its recommended hyperparameter space and report the best performance. 
All experimental details are provided in Appendices~\ref{sec:details} and~\ref{sec:methods}.

\subsection{Experimental Results}

\noindent \textbf{Results on instruction-following}.
As shown in Table~\ref{tab:ae2_mtb}, we report the performance of each model on commonly used instruction-following benchmarks. All preference optimization methods improve over the initial policy, with \RMiPO~achieving the best overall performance without requiring manual hyperparameter tuning. Under the Llama3-8B-Instruct setting, \RMiPO~attains 49.1\% on the LC metric of AlpacaEval 2 and 8.3 on MT-Bench, both surpassing state-of-the-art approaches. In the Mistral-7B-Instruct setting, \RMiPO~achieves an LC win rate of 38.1\%, corresponding to improvements of 11.3 and 6 percentage points over DPO and SimPO, respectively, demonstrating its effectiveness in capturing human preferences. Although $\alpha$-DPO and $\epsilon$-DPO achieve moderate gains through additional control variables or optimized loss functions, they remain sensitive to hyperparameter choices and incur higher training costs. By contrast, \RMiPO~leverages a fully adaptive hyperparameter mechanism to simplify training while maintaining high-quality alignment. Its performance may be slightly lower than SimPER in rare cases, but the gap is minimal. Furthermore, MT-Bench exhibits low discriminative power across methods, likely due to its limited evaluation set and single-instance scoring protocol. Examples of \RMiPO~generated responses provided in Appendix~\ref{sec:exps} further illustrate that its outputs are not only high in quality but also structurally coherent and well-organized.

\setlength{\tabcolsep}{4pt}
\begin{table*}[t]
    \centering
    \small
    \caption{Evaluation results on selected leaderboard benchmarks. Average Rank (Avg. Rank) is the mean of per-task ranks for each method, with lower values indicating better performance. Competition ranking resolves ties.}
    \label{tab:leaderboard}
    \vskip -1em
    \resizebox{1\textwidth}{!}{%
    \begin{tabular}{cl|cccccc|cc}
    \toprule
    \textbf{Model}
     & \textbf{Method} 
     & \textbf{IFEval}
     & \textbf{BBH}
     & \textbf{MATH}
     & \textbf{GPQA}
     & \textbf{MUSR}
     & \textbf{MMLU-PRO}
     & \textbf{Avg. Score}
     & \textbf{Avg. Rank} \\
    \midrule

    \multirow{7}{*}{\parbox[t]{2.2cm}{\centering \textbf{Mistral-7B} \\ \textbf{Instruct}}}  
    & DPO         & 51.76 & 16.88 & 3.10 & 2.46 & 5.75 & 19.43 & 16.56 & 6.33 \\
    & SimPO       & 46.87 & 22.38 & 2.87 & 3.80 & 9.76 & 19.96 & 17.61 & 3.67 \\
    & $\alpha$-DPO & 52.10 & 19.20 & 3.35 & 2.80 & 6.05 & 19.88 & 17.23 & 3.83 \\
    & $\beta$-DPO  & 51.95 & 17.05 & 3.27 & 2.74 & 8.91 & 19.84 & 17.29 & 4.67 \\
    & $\epsilon$-DPO & 52.05 & 17.25 & \textbf{3.50} & 2.85 & 6.88 & 19.75 & 17.05 & 3.83 \\
    & SimPER      & 47.10 & \textbf{22.50} & 2.95 & \textbf{3.91} & 9.47 & \textbf{20.05} & 17.66 & 3.00 \\
    & \RMiPO       & \textbf{52.76} & 22.04 & \textbf{3.50} & 3.67 & \textbf{9.85} & 19.67 & \textbf{18.58} & \textbf{2.50} \\

    \midrule

    \multirow{7}{*}{\parbox[t]{2.2cm}{\centering \textbf{LLama3-8B} \\ \textbf{Instruct}}} 
    & DPO         & 67.57 & 28.51 & 8.46 & 2.91 & 3.93 & 29.61 & 23.50 & 5.67 \\
    & SimPO       & 65.04 & 26.71 & 8.61 & 5.82 & 8.15 & 27.66 & 23.67 & 5.17 \\
    & $\alpha$-DPO & 68.12 & \textbf{29.13} & 8.75 & 4.15 & 5.67 & 29.80 & 24.27 & 3.00 \\
    & $\beta$-DPO  & 67.93 & 28.85 & 8.41 & 3.46 & 6.16 & 28.07 & 23.81 & 5.00 \\
    & $\epsilon$-DPO & 68.15 & 29.01 & 8.94 & 3.27 & \textbf{8.42} & 29.85 & 24.61 & 2.50 \\
    & SimPER      & 65.91 & 27.97 & 8.74 & \textbf{5.91} & 4.15 & 28.67 & 23.56 & 4.67 \\
    & \RMiPO       & \textbf{69.07} & 29.01 & \textbf{8.96} & 3.50 & 8.41 & \textbf{30.46} & \textbf{24.90} & \textbf{1.83} \\

    \bottomrule
    \end{tabular}}
    \vspace{-0.12in}
\end{table*}

\begin{figure}[t]
\centering
\includegraphics[width=\columnwidth]{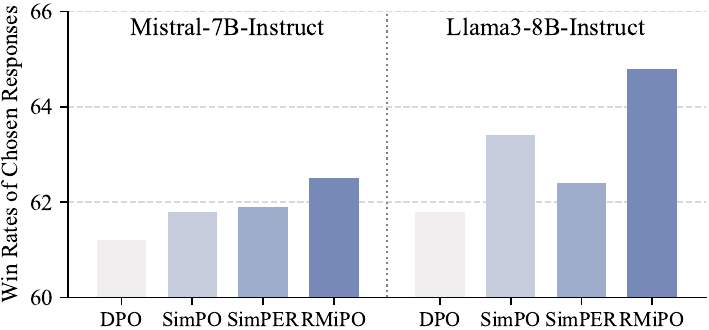} 
\caption{Performance (win rates) of chosen responses on the Anthropic-HH dataset across multiple methods.}
\label{fig:hh}
\vspace{-1.5em}
\end{figure}

\noindent \textbf{Results on downstream tasks}.
To further evaluate the generalization capability of \RMiPO, we conduct a systematic comparative analysis on multiple downstream tasks from the HuggingFace Leaderboard, including IFEval, BBH, MATH, GPQA, MUSR, and MMLU-PRO, with the results summarized in Table~\ref{tab:leaderboard}. Under the Mistral-7B-Instruct setting, \RMiPO~achieves the best average ranking of 2.50, substantially outperforming baseline methods such as DPO (6.33) and SimPO (3.67). This advantage is further amplified under the Llama3-8B-Instruct setting, where \RMiPO~attains an average ranking of 1.83. At the task level, \RMiPO~consistently achieves the highest scores on IFEval across both backbone models (52.76 and 69.07). In addition, \RMiPO~demonstrates strong performance on MATH, with scores of 3.50 and 8.96, highlighting its effectiveness in structured reasoning and mathematical problem solving. Although methods such as $\epsilon$-DPO yield competitive results on specific tasks (e.g., MUSR or GPQA) by employing more complex optimization strategies, their performance varies considerably across tasks. In contrast, \RMiPO~achieves consistently strong results across diverse downstream tasks without incurring additional parameter search costs, underscoring its superior stability and generalization ability.

\noindent \textbf{Results on safety alignment}.
The above benchmarks primarily target open-domain tasks. To evaluate \RMiPO’s performance in terms of safety and preference alignment, we compare it with baselines on the Anthropic-HH dataset. This dataset comprises the Helpful and Harmless subsets, with alignment effectiveness measured via the win rate of chosen responses. As shown in Figure~\ref{fig:hh}, \RMiPO~achieves significantly higher average win rates across both subsets compared with DPO, SimPO, and its improved variant SimPER. Under the Llama3-8B-Instruct setting, \RMiPO~attains an average win rate of 64.8\%, surpassing DPO’s 61.8\%. Notably, on the Harmless subset, \RMiPO~demonstrates even greater gains, while maintaining a stable lead without requiring complex hyperparameter tuning on safety-related data, highlighting its robustness and practical utility. 

\begin{table}[t]
\centering

\caption{Ablation analysis on leaderboard benchmarks. GM: tuning $\gamma$; MI: intrinsic mutual information–based fine-tuning; ED: exponential decay–based update.}
\label{tab:ablation}
\vskip -0.8em
\setlength{\tabcolsep}{2pt}
\resizebox{\columnwidth}{!}{%
\begin{tabular}{ccc|cccccc}
\toprule
GM & MI & ED & IFEval & BBH & MATH & GPQA & MUSR & MMLU-PRO \\ 
\midrule
\faCheck & & & 66.14 & 27.32 & 8.52 & 3.01 & 4.25 & 28.04 \\
 & \faCheck & & 65.87 & 26.94 & 8.48 & 3.12 & 5.12 & 28.32 \\
 \faCheck & \faCheck & & 68.21 & 28.47 & 8.74 & 3.28 & 7.31 & 29.56 \\
 \faCheck & & \faCheck & 67.58 & 27.82 & 8.61 & 3.07 & 6.47 & 28.91 \\
 & \faCheck & \faCheck & 68.82 & \textbf{29.73} & 8.85 & \textbf{4.18} & 8.28 & 29.87 \\
 \faCheck & \faCheck & \faCheck & \textbf{69.07} & 29.01 & \textbf{8.96} & 3.50 & \textbf{8.41} & \textbf{30.46} \\
\bottomrule
\end{tabular}}
\vspace{-1em}
\end{table}

\noindent \textbf{Ablation analysis}.
We conduct an ablation study of the three key design components in \RMiPO on the Llama3-8B-Instruct model, with the results summarized in Table~\ref{tab:ablation}. When only one of GM or MI is enabled, the model achieves relatively limited performance across benchmarks (e.g., IFEval scores of 66.14 and 65.87, respectively), indicating that a single component cannot fully unlock the model’s potential. As components are progressively combined, overall performance improves substantially. For example, combining GM and MI increases MMLU-PRO from approximately 28\% to 29.56, while the combination of MI and ED performs particularly well on BBH (29.73) and GPQA (4.18). When all three components (GM + MI + ED) are enabled simultaneously, the model achieves the best results on core benchmarks including IFEval (69.07), MATH (8.96), MUSR (8.41), and MMLU-PRO (30.46). These results indicate that the performance gains primarily stem from the synergistic interactions among \RMiPO’s components, with the full system delivering the best overall performance.

\begin{figure}[t]
\centering
\includegraphics[width=\columnwidth]{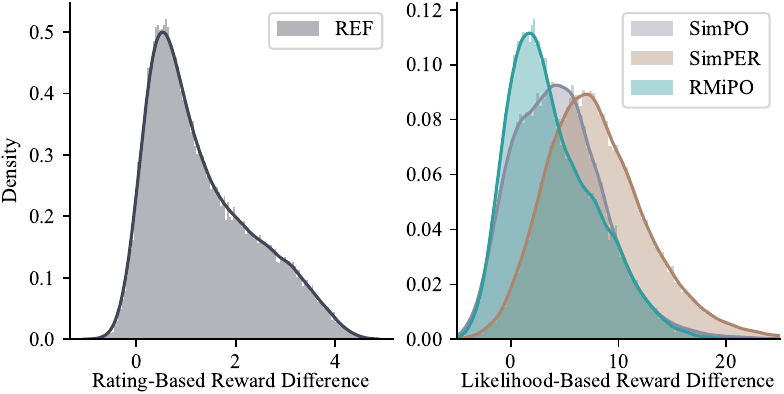} 
\caption{Density analysis of reward differences, with REF serving as the ground-truth reference.}
\label{fig:density}
\vspace{-1.5em}
\end{figure}

\noindent \textbf{Density analysis}.
We analyze the reward density of the Llama3-8B-Instruct model. Specifically, we construct reward differences from the rankings of responses in the UltraFeedback dataset, where each difference is defined as the ranking of the preferred response minus that of the less-preferred response. Since the ranking-based reward differences are discrete, we apply bounded uniform jittering to make them continuous, thereby approximating the latent reward. This strategy is justified because the original rankings are generated by a higher-quality oracle model, which serves as a reliable proxy for human preferences. As shown in Figure~\ref{fig:density}, we compare the performance of \RMiPO~with that of SimPO and SimPER. The results indicate that the reward distribution generated by \RMiPO~aligns most closely with the reference benchmark distribution. This demonstrates its ability to capture fine-grained differences in human preference intensity and provides strong evidence for the superiority of \RMiPO.

\begin{figure}[t]
\centering
\includegraphics[width=\columnwidth]{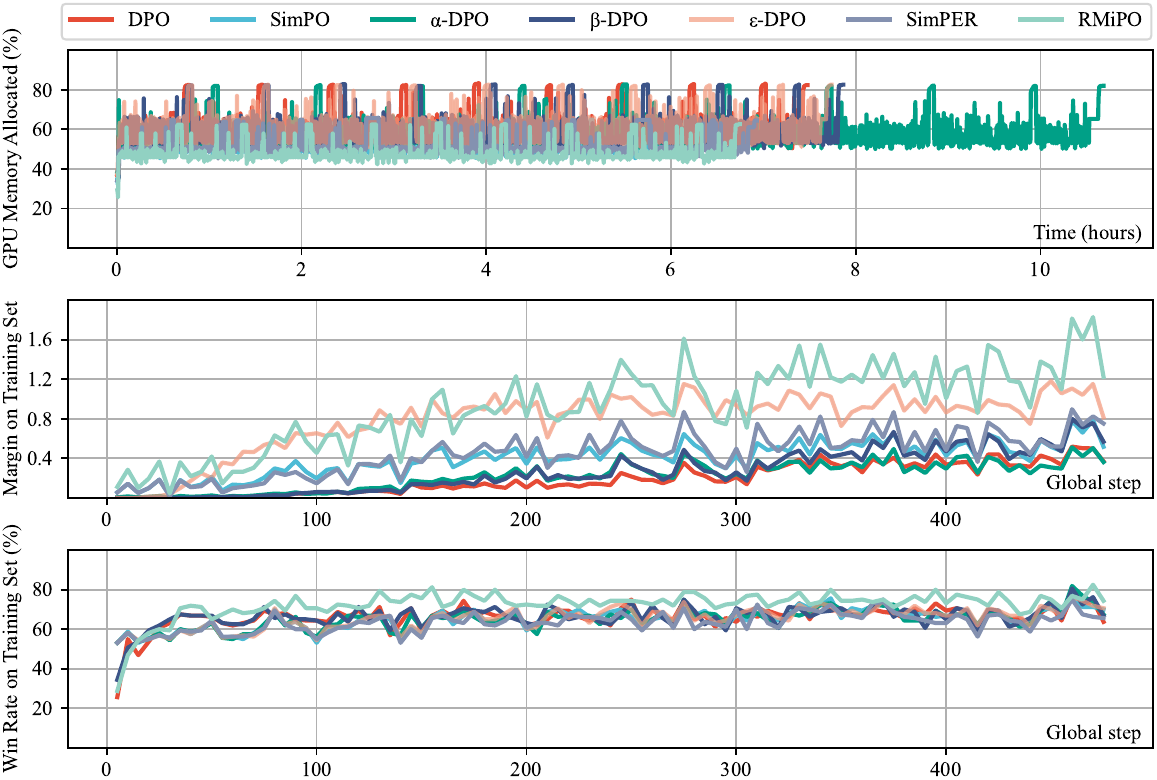} 
\caption{Training efficiency and dynamics, showing cost, margin, and win rate on the training set.}
\label{fig:time}
\vspace{-1.5em}
\end{figure}

\noindent \textbf{Efficiency analysis}.
To further evaluate the practicality of each algorithm in real-world production environments, we analyzed the computational efficiency and training stability of \RMiPO~and its baseline methods, focusing on GPU memory usage, training duration, and performance convergence. As shown in Figure~\ref{fig:time}, \RMiPO~demonstrates superior efficiency, requiring only approximately 6.75 hours of training with a peak GPU memory usage of about 62\%. In comparison, SimPO and SimPER require 6.17 hours and 6.92 hours, respectively, with similar peak memory usage, while DPO and its variants $\beta$-DPO and $\epsilon$-DPO exhibit peak memory usage as high as 82\% and training durations approaching 8 hours. Notably, $\alpha$-DPO requires approximately 10 hours of training due to complex dynamic reference model computations and additional preprocessing steps. During training, \RMiPO~consistently outperforms other methods in Margin and Win Rate metrics on the training set, maintaining a leading position throughout the entire training cycle. Overall, \RMiPO~combines high efficiency, low resource consumption, and superior performance, making it a highly practical solution for large-scale language model alignment tasks.


\section{Related Work}

\noindent \textbf{Preference Optimization for LLM}. Preference optimization is a key technique for aligning LLMs with human values. This paradigm was originally proposed by~\cite{ouyang2022training} and employs proximal policy optimization for online training. However, the high complexity of online algorithms has driven researchers to shift toward more efficient offline methods. Offline preference optimization~\cite{park2024disentangling, rafailov2023direct, zhao2023slic, azar2024general, ethayarajh2024kto, xu2024contrastive, zeng2024token, amini2024direct, tanggeneralized, liu2025lipo, liu2025understanding} directly leverages paired preference data to guide policies toward preferred behaviors. Building on this line of work, ~\cite{meng2024simpo, hong2024orpo} proposed an approach that does not require a reference model. Despite their effectiveness, these methods require extensive hyperparameter tuning.

\noindent \textbf{Efficiency Improvement in DPO}. Unlike prior work on data filtering~\cite{muldrew2024active, yang2024not, huanglarger25, corrado2025automixalign}, this study focuses on minimizing hyperparameter tuning overhead in DPO and its variants. Existing research has explored several strategies: ~\cite{wu2024beta} and~\cite{lee2025kl} dynamically adjust hyperparameters via KL regularization to account for instance-level preference intensity; ~\cite{wu2025alphadpo, sun2025robust} introduces an implicit reference model, reducing hyperparameters that require tuning; and ~\cite{xiao2025simper} proposes the first parameter-free preference optimization objective based on perplexity. While these methods demonstrate success, they remain limited in effectiveness. To address this, we propose \RMiPO, an adaptive hyperparameter tuning mechanism based on intrinsic mutual information that improves alignment while reducing optimization overhead.


\section{Conclusion}

In this paper, we propose \RMiPO, a lightweight and efficient offline preference optimization framework. Its core innovation is an adaptive modulation mechanism based on intrinsic response-level mutual information. This mechanism dynamically decouples the contribution of different samples to the overall preference signal, enabling finer-grained alignment control. Theoretical analysis shows that this mutual information–driven strategy offers strong interpretability, establishing a new principled paradigm for sample weighting in preference optimization. Extensive experiments demonstrate that \RMiPO~achieves superior alignment gains with lower training costs and favorable convergence.


\section*{Limitations}

The core advantage of \RMiPO~lies in its ability to obviate the manual tuning of task-sensitive hyperparameters, rather than reducing the number of tunable parameters. This property substantially enhances the practicality of preference optimization algorithms. Nevertheless, \RMiPO~still exhibits several limitations that warrant further investigation.

\noindent\textbf{Sensitivity}. While $\gamma$ is the primary driver of adaptivity, its efficacy presupposes that $\beta$ lies within a reasonable range, as typically ensured by standard model-scaling practices. Although extreme $\beta$ values destabilize all preference-optimization baselines, \RMiPO~exhibits greater robustness, with performance degrading more gracefully than static-margin methods. By dynamically adjusting the reward margin $\gamma$ (and its bounds) as a function of $\beta$, the proposed exponential decay mechanism partially mitigates mis-scaled preference signals and reduces performance degradation under suboptimal settings. Although \RMiPO~cannot compensate for arbitrarily ill-chosen $\beta$, its adaptive margin consistently offers a safety buffer over fixed-margin baselines. The coupled dynamics between $\beta$ and $\gamma$ therefore warrant further study in the context of optimization stability and generalization.

\noindent\textbf{Scalability}. Recent studies suggest increasing policy optimization iterations can further improve instruction-following performance. However, due to computational constraints, this work, like most existing studies, does not conduct extensive validation under large-scale regimes. Our experiments focus on billion-parameter models, which suffice to elucidate \RMiPO’s key mechanisms and provide empirical support for its effectiveness. The generalization of \RMiPO~to ten-billion-parameter models remains to be systematically evaluated.

\noindent\textbf{Adaptability}. This work focuses on offline preference optimization with fixed datasets, which limits adaptation to dynamic environments. Extending \RMiPO~to an online learning framework with real-time interaction feedback could improve its applicability to scenarios such as human–computer interaction and autonomous agents, for example by using an online oracle to provide preferences and intrinsic mutual information to guide exploration. 



\section*{Ethical Considerations}

This work proposes a lightweight framework for LLM alignment. The study does not involve human participants or personal or sensitive data, nor does it enable malicious use; therefore, it raises no ethical concerns. The proposed method aims to improve the robustness and efficiency of value alignment, supporting the reliable deployment of large language models. Furthermore, LLM-based assistants were employed solely for linguistic refinement, with the authors explicitly retaining full responsibility for the research design, technical content, and final conclusions.

\section*{Acknowledgments}

This work was supported by the Science and Technology Development Fund of the Macau SAR (Grant No. 0119/2025/RIA2) and the National Natural Science Foundation of China (Grant Nos. W2512008, 62572500, and 62272498).


\bibliography{custom}

@misc{llama2025,
  author       = {Llama},
  title        = {The Llama 4 herd: The beginning of a new era of natively multimodal AI innovation},
  year         = {2025},
  howpublished = {\url{https://ai.meta.com/blog/llama-4-multimodal-intelligence/}},
  note         = {Accessed: 2025-10-8},
}

@misc{deepseek2025,
  author       = {DeepSeek},
  title        = {Introducing DeepSeek-V3.2-Exp},
  year         = {2025},
  howpublished = {\url{https://api-docs.deepseek.com/news/news250929}},
  note         = {Accessed: 2025-10-8},
}

@misc{gpt2025,
    author = {OpenAI},
    title = {Introducing GPT-5},
    howpublished = {\url{https://openai.com/index/introducing-gpt-5/}},
    year = {2025},
    note = {Accessed: 2025-10-8},
}

@article{ouyang2022training,
  title={Training language models to follow instructions with human feedback},
  author = {Ouyang, Long and Wu, Jeff and Jiang, Xu and Almeida, Diogo and Wainwright, Carroll L. and Mishkin, Pamela and Zhang, Chong and Agarwal, Sandhini and Slama, Katarina and Ray, Alex and Schulman, John and Hilton, Jacob and Kelton, Fraser and Miller, Luke and Simens, Maddie and Askell, Amanda and Welinder, Peter and Christiano, Paul and Leike, Jan and Lowe, Ryan},
  journal={Advances in neural information processing systems},
  volume={35},
  pages={27730--27744},
  year={2022}
}

@article{rafailov2023direct,
  title={Direct preference optimization: Your language model is secretly a reward model},
  author={Rafailov, Rafael and Sharma, Archit and Mitchell, Eric and Manning, Christopher D and Ermon, Stefano and Finn, Chelsea},
  journal={Advances in neural information processing systems},
  volume={36},
  pages={53728--53741},
  year={2023}
}

@article{ethayarajh2024kto,
  title={Kto: Model alignment as prospect theoretic optimization},
  author={Ethayarajh, Kawin and Xu, Winnie and Muennighoff, Niklas and Jurafsky, Dan and Kiela, Douwe},
  journal={arXiv preprint arXiv:2402.01306},
  year={2024}
}

@inproceedings{azar2024general,
  title={A general theoretical paradigm to understand learning from human preferences},
  author={Azar, Mohammad Gheshlaghi and Guo, Zhaohan Daniel and Piot, Bilal and Munos, Remi and Rowland, Mark and Valko, Michal and Calandriello, Daniele},
  booktitle={International Conference on Artificial Intelligence and Statistics},
  pages={4447--4455},
  year={2024},
  organization={PMLR}
}

@misc{huggingface2024_ft,
    author = {Rasul, Kashif and Beeching, Edward and Tunstall, Lewis and von Werra, Leandro and Sanseviero, Omar},
    title = {Preference Tuning LLMs with Direct Preference Optimization Methods},
    howpublished = {\url{https://huggingface.co/blog/pref-tuning}},
    year = {2024},
    note = {Accessed: 2024-11-27},
}

@article{meng2024simpo,
  title={Simpo: Simple preference optimization with a reference-free reward},
  author={Meng, Yu and Xia, Mengzhou and Chen, Danqi},
  journal={Advances in Neural Information Processing Systems},
  volume={37},
  pages={124198--124235},
  year={2024}
}

@inproceedings{wu2024beta,
  title={$\beta$-DPO: direct preference optimization with dynamic $\beta$},
  author={Wu, Junkang and Xie, Yuexiang and Yang, Zhengyi and Wu, Jiancan and Gao, Jinyang and Ding, Bolin and Wang, Xiang and He, Xiangnan},
  booktitle={Proceedings of the 38th International Conference on Neural Information Processing Systems},
  pages={129944--129966},
  year={2024}
}

@article{lee2025kl,
  title={KL Penalty Control via Perturbation for Direct Preference Optimization},
  author={Lee, Sangkyu and Han, Janghoon and Song, Hosung and Choi, Stanley Jungkyu and Lee, Honglak and Yu, Youngjae},
  journal={arXiv preprint arXiv:2502.13177},
  year={2025}
}

@misc{wu2025alphadpo,
      title={AlphaDPO: Adaptive Reward Margin for Direct Preference Optimization}, 
      author={Junkang Wu and Xue Wang and Zhengyi Yang and Jiancan Wu and Jinyang Gao and Bolin Ding and Xiang Wang and Xiangnan He},
      year={2025},
      eprint={2410.10148},
      archivePrefix={arXiv},
      primaryClass={cs.LG},
      url={https://arxiv.org/abs/2410.10148}, 
}

@inproceedings{li2024contrastive,
  title={Contrastive continual learning with importance sampling and prototype-instance relation distillation},
  author={Li, Jiyong and Azizov, Dilshod and Liang, Shangsong and others},
  booktitle={Proceedings of the AAAI Conference on Artificial Intelligence},
  volume={38},
  pages={13554--13562},
  year={2024}
}

@article{xiao2025simper,
  title={SimPER: A minimalist approach to preference alignment without hyperparameters},
  author={Xiao, Teng and Yuan, Yige and Chen, Zhengyu and Li, Mingxiao and Liang, Shangsong and Ren, Zhaochun and Honavar, Vasant G},
  journal={arXiv preprint arXiv:2502.00883},
  year={2025}
}

@article{yang2024not,
  title={Not all preference pairs are created equal: A recipe for annotation-efficient iterative preference learning},
  author={Yang, Sen and Cui, Leyang and Cai, Deng and Huang, Xinting and Shi, Shuming and Lam, Wai},
  journal={arXiv preprint arXiv:2406.17312},
  year={2024}
}

@article{gorbatovski2024learn,
  title={Learn your reference model for real good alignment},
  author={Gorbatovski, Alexey and Shaposhnikov, Boris and Malakhov, Alexey and Surnachev, Nikita and Aksenov, Yaroslav and Maksimov, Ian and Balagansky, Nikita and Gavrilov, Daniil},
  journal={arXiv preprint arXiv:2404.09656},
  year={2024}
}

@article{zeng2024token,
  title={Token-level direct preference optimization},
  author={Zeng, Yongcheng and Liu, Guoqing and Ma, Weiyu and Yang, Ning and Zhang, Haifeng and Wang, Jun},
  journal={arXiv preprint arXiv:2404.11999},
  year={2024}
}

@article{christiano2017deep,
  title={Deep reinforcement learning from human preferences},
  author={Christiano, Paul F and Leike, Jan and Brown, Tom and Martic, Miljan and Legg, Shane and Amodei, Dario},
  journal={Advances in neural information processing systems},
  volume={30},
  year={2017}
}

@article{bradley1952rank,
  title={Rank analysis of incomplete block designs: I. the method of paired comparisons},
  author={Bradley, Ralph Allan and Terry, Milton E},
  journal={Biometrika},
  volume={39},
  number={3/4},
  pages={324--345},
  year={1952},
  publisher={JSTOR}
}

@article{zheng2023judging,
  title={Judging llm-as-a-judge with mt-bench and chatbot arena},
  author = {Zheng, Lianmin and Chiang, Wei-Lin and Sheng, Ying and Zhuang, Siyuan and Wu, Zhanghao and Zhuang, Yonghao and Lin, Zi and Li, Zhuohan and Li, Dacheng and Xing, Eric P. and Zhang, Hao and Gonzalez, Joseph E. and Stoica, Ion},
  journal={Advances in neural information processing systems},
  volume={36},
  pages={46595--46623},
  year={2023}
}

@article{bai2022training,
  title={Training a helpful and harmless assistant with reinforcement learning from human feedback},
  author = {Bai, Yuntao and Jones, Andy and Ndousse, Kamal and Askell, Amanda and Chen, Anna and DasSarma, Nova and Drain, Dawn and Fort, Stanislav and Ganguli, Deep and Henighan, Tom and Joseph, Nicholas and Kadavath, Saurav and Kernion, Jackson and Conerly, Tom and El-Showk, Sheer and Elhage, Nelson and Hatfield-Dodds, Zac and Hernandez, Danny and Hume, Tristan and Johnston, Scott and Kravec, Shauna and Lovitt, Liane and Nanda, Neel and Olsson, Catherine and Amodei, Dario and Brown, Tom and Clark, Jack and McCandlish, Sam and Olah, Chris and Mann, Ben and Kaplan, Jared},
  journal={arXiv preprint arXiv:2204.05862},
  year={2022}
}

@inproceedings{liu2025pointwise,
  title={Pointwise mutual information as a performance gauge for retrieval-augmented generation},
  author={Liu, Tianyu and Qi, Jirui and He, Paul and Bisazza, Arianna and Sachan, Mrinmaya and Cotterell, Ryan},
  booktitle={Proceedings of the 2025 Conference of the Nations of the Americas Chapter of the Association for Computational Linguistics: Human Language Technologies (Volume 1: Long Papers)},
  pages={1628--1647},
  year={2025}
}

@inproceedings{chen2024language,
  title={Language model based unsupervised dependency parsing with conditional mutual information and grammatical constraints},
  author={Chen, Junjie and He, Xiangheng and Miyao, Yusuke},
  booktitle={Proceedings of the 2024 Conference of the North American Chapter of the Association for Computational Linguistics: Human Language Technologies (Volume 1: Long Papers)},
  pages={6355--6366},
  year={2024}
}

@inproceedings{li2016diversity,
  title={A diversity-promoting objective function for neural conversation models},
  author={Li, Jiwei and Galley, Michel and Brockett, Chris and Gao, Jianfeng and Dolan, William B},
  booktitle={Proceedings of the 2016 conference of the North American chapter of the association for computational linguistics: human language technologies},
  pages={110--119},
  year={2016}
}

@article{dubey2024llama,
  title={The llama 3 herd of models},
  author = {Grattafiori, Aaron and Dubey, Abhimanyu and Jauhri, Abhinav and Pandey, Abhinav and Kadian, Abhishek and Al-Dahle, Ahmad and Letman, Aiesha and Mathur, Akhil and Schelten, Alan and Vaughan, Alex and Yang, Amy and Fan, Angela and Goyal, Anirudh and Hartshorn, Anthony and Yang, Aobo and Mitra, Archi and Sravankumar, Archie and Korenev, Artem and Hinsvark, Arthur and Rao, Arun and Zhang, Aston and Rodriguez, Aurelien and Gregerson, Austen and Spataru, Ava and Roziere, Baptiste and Biron, Bethany and Tang, Binh and Chern, Bobbie and Caucheteux, Charlotte and Nayak, Chaya and Bi, Chloe and Marra, Chris and McConnell, Chris and Keller, Christian and Touret, Christophe and Wu, Chunyang and Wong, Corinne and Ferrer, Cristian Canton and Nikolaidis, Cyrus and Allonsius, Damien and Song, Daniel and Pintz, Danielle and Livshits, Danny and Wyatt, Danny and Esiobu, David and Choudhary, Dhruv and Mahajan, Dhruv and Garcia-Olano, Diego and Perino, Diego and Hupkes, Dieuwke and Lakomkin, Egor and AlBadawy, Ehab and Lobanova, Elina and Dinan, Emily and Smith, Eric Michael and Radenovic, Filip and Guzmán, Francisco and Zhang, Frank and Synnaeve, Gabriel and Lee, Gabrielle and Anderson, Georgia Lewis and Thattai, Govind and Nail, Graeme and Mialon, Gregoire and Pang, Guan and Cucurell, Guillem and Nguyen, Hailey and Korevaar, Hannah and Xu, Hu and Touvron, Hugo and Zarov, Iliyan and Ibarra, Imanol Arrieta and Kloumann, Isabel and Misra, Ishan and Evtimov, Ivan and Zhang, Jack and Copet, Jade and Lee, Jaewon and Geffert, Jan and Vranes, Jana and Park, Jason and Mahadeokar, Jay and Shah, Jeet and van der Linde, Jelmer and Billock, Jennifer and Hong, Jenny and Lee, Jenya and Fu, Jeremy and Chi, Jianfeng and Huang, Jianyu and Liu, Jiawen and Wang, Jie and Yu, Jiecao and Bitton, Joanna and Spisak, Joe and Park, Jongsoo and Rocca, Joseph and Johnstun, Joshua and Saxe, Joshua and Jia, Junteng and Alwala, Kalyan Vasuden and Prasad, Karthik and Upasani, Kartikeya and Plawiak, Kate and Li, Ke and Heafield, Kenneth and Stone, Kevin and El-Arini, Khalid and Iyer, Krithika and Malik, Kshitiz and Chiu, Kuenley and Bhalla, Kunal and Lakhotia, Kushal and Rantala-Yeary, Lauren and van der Maaten, Laurens and Chen, Lawrence and Tan, Liang and Jenkins, Liz and Martin, Louis and Madaan, Lovish and Malo, Lubo and Blecher, Lukas and Landzaat, Lukas and de Oliveira, Luke and Muzzi, Madeline and Pasupuleti, Mahesh and Singh, Mannat and Paluri, Manohar and Kardas, Marcin and Tsimpoukelli, Maria and Oldham, Mathew and Rita, Mathieu and Pavlova, Maya and Kambadur, Melanie and Lewis, Mike and Si, Min and Singh, Mitesh Kumar and Hassan, Mona and Goyal, Naman and Torabi, Narjes and Bashlykov, Nikolay and Bogoychev, Nikolay and Chatterji, Niladri and Zhang, Ning and Duchenne, Olivier and Çelebi, Onur and Alrassy, Patrick and Zhang, Pengchuan and Li, Pengwei and Vasic, Petar and Weng, Peter and Bhargava, Prajjwal and Dubal, Pratik and Krishnan, Praveen and Koura, Punit Singh and Xu, Puxin and He, Qing and Dong, Qingxiao and Srinivasan, Ragavan and Ganapathy, Raj and Calderer, Ramon and Cabral, Ricardo Silveira and Stojnic, Robert and Raileanu, Roberta and Maheswari, Rohan and Girdhar, Rohit and Patel, Rohit and Sauvestre, Romain and Polidoro, Ronnie and Sumbaly, Roshan and Taylor, Ross and Silva, Ruan and Hou, Rui and Wang, Rui and Hosseini, Saghar and Chennabasappa, Sahana and Singh, Sanjay and Bell, Sean and Kim, Seohyun Sonia and Edunov, Sergey and Nie, Shaoliang and Narang, Sharan and Raparthy, Sharath and Shen, Sheng and Wan, Shengye and Bhosale, Shruti and Zhang, Shun and Vandenhende, Simon and Batra, Soumya and Whitman, Spencer and Sootla, Sten and Collot, Stephane and Gururangan, Suchin and Borodinsky, Sydney and Herman, Tamar and Fowler, Tara and Sheasha, Tarek and Georgiou, Thomas Brandon and Scialom, Thomas and Speckbacher, Tobias and Mihaylov, Todor and Xiao, Tong and Karn, Ujjwal and Goswami, Vedanuj and Gupta, Vibhor and Ramanathan, Vignesh and Kerkez, Viktor and Gonguet, Vincent and Do, Virginie and Vogeti, Vish and Albiero, Vítor and Petrovic, Vladan and Chu, Weiwei and Xiong, Wenhan and Fu, Wenyin and Meers, Whitney and Martinet, Xavier and Wang, Xiaodong and Wang, Xiaofang and Tan, Xiaoqing Ellen and Xia, Xide and Xie, Xinfeng and Jia, Xuchao and Wang, Xuewei and Goldschlag, Yaelle and Gaur, Yashesh and Babaei, Yasmine and Wen, Yi and Song, Yiwen and Zhang, Yuchen and Li, Yue and Mao, Yuning and Coudert, Zacharie Delpierre and Yan, Zheng and Chen, Zhengxing and Papakipos, Zoe and Singh, Aaditya and Srivastava, Aayushi and Jain, Abha and Kelsey, Adam and Shajnfeld, Adam and Gangidi, Adithya and Victoria, Adolfo and Goldstand, Ahuva and Menon, Ajay and Sharma, Ajay and Boesenberg, Alex and Baevski, Alexei and Feinstein, Allie and Kallet, Amanda and Sangani, Amit and Teo, Amos and Yunus, Anam and Lupu, Andrei and Alvarado, Andres and Caples, Andrew and Gu, Andrew and Ho, Andrew and Poulton, Andrew and Ryan, Andrew and Ramchandani, Ankit and Dong, Annie and Franco, Annie and Goyal, Anuj and Saraf, Aparajita and Chowdhury, Arkabandhu and Gabriel, Ashley and Bharambe, Ashwin and Eisenman, Assaf and Yazdan, Azadeh and James, Beau and Maurer, Ben and Leonhardi, Benjamin and Huang, Bernie and Loyd, Beth and De Paola, Beto and Paranjape, Bhargavi and Liu, Bing and Wu, Bo and Ni, Boyu and Hancock, Braden and Wasti, Bram and Spence, Brandon and Stojkovic, Brani and Gamido, Brian and Montalvo, Britt and Parker, Carl and Burton, Carly and Mejia, Catalina and Liu, Ce and Wang, Changhan and Kim, Changkyu and Zhou, Chao and Hu, Chester and Chu, Ching-Hsiang and Cai, Chris and Tindal, Chris and Feichtenhofer, Christoph and Gao, Cynthia and Civin, Damon and Beaty, Dana and Kreymer, Daniel and Li, Daniel and Adkins, David and Xu, David and Testuggine, Davide and David, Delia and Parikh, Devi and Liskovich, Diana and Foss, Didem and Wang, Dingkang and Le, Duc and Holland, Dustin and Dowling, Edward and Jamil, Eissa and Montgomery, Elaine and Presani, Eleonora and Hahn, Emily and Wood, Emily and Le, Eric-Tuan and Brinkman, Erik and Arcaute, Esteban and Dunbar, Evan and Smothers, Evan and Sun, Fei and Kreuk, Felix and Tian, Feng and Kokkinos, Filippos and Ozgenel, Firat and Caggioni, Francesco and Kanayet, Frank and Seide, Frank and Florez, Gabriela Medina and Schwarz, Gabriella and Badeer, Gada and Swee, Georgia and Halpern, Gil and Herman, Grant and Sizov, Grigory and Zhang, Guangyi and Lakshminarayanan, Guna and Inan, Hakan and Shojanazeri, Hamid and Zou, Han and Wang, Hannah and Zha, Hanwen and Habeeb, Haroun and Rudolph, Harrison and Suk, Helen and Aspegren, Henry and Goldman, Hunter and Zhan, Hongyuan and Damlaj, Ibrahim and Molybog, Igor and Tufanov, Igor and Leontiadis, Ilias and Veliche, Irina-Elena and Gat, Itai and Weissman, Jake and Geboski, James and Kohli, James and Lam, Janice and Asher, Japhet and Gaya, Jean-Baptiste and Marcus, Jeff and Tang, Jeff and Chan, Jennifer and Zhen, Jenny and Reizenstein, Jeremy and Teboul, Jeremy and Zhong, Jessica and Jin, Jian and Yang, Jingyi and Cummings, Joe and Carvill, Jon and Shepard, Jon and McPhie, Jonathan and Torres, Jonathan and Ginsburg, Josh and Wang, Junjie and Wu, Kai and U, Kam Hou and Saxena, Karan and Khandelwal, Kartikay and Zand, Katayoun and Matosich, Kathy and Veeraraghavan, Kaushik and Michelena, Kelly and Li, Keqian and Jagadeesh, Kiran and Huang, Kun and Chawla, Kunal and Huang, Kyle and Chen, Lailin and Garg, Lakshya and A, Lavender and Silva, Leandro and Bell, Lee and Zhang, Lei and Guo, Liangpeng and Yu, Licheng and Moshkovich, Liron and Wehrstedt, Luca and Khabsa, Madian and Avalani, Manav and Bhatt, Manish and Mankus, Martynas and Hasson, Matan and Lennie, Matthew and Reso, Matthias and Groshev, Maxim and Naumov, Maxim and Lathi, Maya and Keneally, Meghan and Liu, Miao and Seltzer, Michael L. and Valko, Michal and Restrepo, Michelle and Patel, Mihir and Vyatskov, Mik and Samvelyan, Mikayel and Clark, Mike and Macey, Mike and Wang, Mike and Hermoso, Miquel Jubert and Metanat, Mo and Rastegari, Mohammad and Bansal, Munish and Santhanam, Nandhini and Parks, Natascha and White, Natasha and Bawa, Navyata and Singhal, Nayan and Egebo, Nick and Usunier, Nicolas and Mehta, Nikhil and Laptev, Nikolay Pavlovich and Dong, Ning and Cheng, Norman and Chernoguz, Oleg and Hart, Olivia and Salpekar, Omkar and Kalinli, Ozlem and Kent, Parkin and Parekh, Parth and Saab, Paul and Balaji, Pavan and Rittner, Pedro and Bontrager, Philip and Roux, Pierre and Dollar, Piotr and Zvyagina, Polina and Ratanchandani, Prashant and Yuvraj, Pritish and Liang, Qian and Alao, Rachad and Rodriguez, Rachel and Ayub, Rafi and Murthy, Raghotham and Nayani, Raghu and Mitra, Rahul and Parthasarathy, Rangaprabhu and Li, Raymond and Hogan, Rebekkah and Battey, Robin and Wang, Rocky and Howes, Russ and Rinott, Ruty and Mehta, Sachin and Siby, Sachin and Bondu, Sai Jayesh and Datta, Samyak and Chugh, Sara and Hunt, Sara and Dhillon, Sargun and Sidorov, Sasha and Pan, Satadru and Mahajan, Saurabh and Verma, Saurabh and Yamamoto, Seiji and Ramaswamy, Sharadh and Lindsay, Shaun and Feng, Sheng and Lin, Shenghao and Zha, Shengxin Cindy and Patil, Shishir and Shankar, Shiva and Zhang, Shuqiang and Wang, Sinong and Agarwal, Sneha and Sajuyigbe, Soji and Chintala, Soumith and Max, Stephanie and Chen, Stephen and Kehoe, Steve and Satterfield, Steve and Govindaprasad, Sudarshan and Gupta, Sumit and Deng, Summer and Cho, Sungmin and Virk, Sunny and Subramanian, Suraj and Choudhury, Sy and Goldman, Sydney and Remez, Tal and Glaser, Tamar and Best, Tamara and Koehler, Thilo and Robinson, Thomas and Li, Tianhe and Zhang, Tianjun and Matthews, Tim and Chou, Timothy and Shaked, Tzook and Vontimitta, Varun and Ajayi, Victoria and Montanez, Victoria and Mohan, Vijai and Kumar, Vinay Satish and Mangla, Vishal and Ionescu, Vlad and Poenaru, Vlad and Mihailescu, Vlad Tiberiu and Ivanov, Vladimir and Li, Wei and Wang, Wenchen and Jiang, Wenwen and Bouaziz, Wes and Constable, Will and Tang, Xiaocheng and Wu, Xiaojian and Wang, Xiaolan and Wu, Xilun and Gao, Xinbo and Kleinman, Yaniv and Chen, Yanjun and Hu, Ye and Jia, Ye and Qi, Ye and Li, Yenda and Zhang, Yilin and Zhang, Ying and Adi, Yossi and Nam, Youngjin and Wang, Yu and Zhao, Yu and Hao, Yuchen and Qian, Yundi and Li, Yunlu and He, Yuzi and Rait, Zach and DeVito, Zachary and Rosnbrick, Zef and Wen, Zhaoduo and Yang, Zhenyu and Zhao, Zhiwei and Ma, Zhiyu},
  journal={arXiv e-prints},
  pages={arXiv--2407},
  year={2024}
}

@misc{jiang2023mistral7b,
      title={Mistral 7B}, 
      author={Albert Q. Jiang and Alexandre Sablayrolles and Arthur Mensch and Chris Bamford and Devendra Singh Chaplot and Diego de las Casas and Florian Bressand and Gianna Lengyel and Guillaume Lample and Lucile Saulnier and Lélio Renard Lavaud and Marie-Anne Lachaux and Pierre Stock and Teven Le Scao and Thibaut Lavril and Thomas Wang and Timothée Lacroix and William El Sayed},
      year={2023},
      eprint={2310.06825},
      archivePrefix={arXiv},
      primaryClass={cs.CL},
      url={https://arxiv.org/abs/2310.06825}, 
}

@article{tunstall2023zephyr,
  title={Zephyr: Direct distillation of lm alignment},
  author = {Tunstall, Lewis and Beeching, Edward and Lambert, Nathan and Rajani, Nazneen and Rasul, Kashif and Belkada, Younes and Huang, Shengyi and von Werra, Leandro and Fourrier, Clémentine and Habib, Nathan and Sarrazin, Nathan and Sanseviero, Omar and Rush, Alexander M. and Wolf, Thomas},
  journal={arXiv preprint arXiv:2310.16944},
  year={2023}
}

@misc{cui2023ultrafeedback,
      title={UltraFeedback: Boosting Language Models with Scaled AI Feedback}, 
      author={Ganqu Cui and Lifan Yuan and Ning Ding and Guanming Yao and Bingxiang He and Wei Zhu and Yuan Ni and Guotong Xie and Ruobing Xie and Yankai Lin and Zhiyuan Liu and Maosong Sun},
      year={2024},
      eprint={2310.01377},
      archivePrefix={arXiv},
      primaryClass={cs.CL},
      url={https://arxiv.org/abs/2310.01377}, 
}

@misc{alpaca_eval,
  author = {Xuechen Li and Tianyi Zhang and Yann Dubois and Rohan Taori and Ishaan Gulrajani and Carlos Guestrin and Percy Liang and Tatsunori B. Hashimoto },
  title = {AlpacaEval: An Automatic Evaluator of Instruction-following Models},
  year = {2023},
  month = {5},
  publisher = {GitHub},
  journal = {GitHub repository},
  howpublished = {\url{https://github.com/tatsu-lab/alpaca_eval}}
}

@article{zhao2023slic,
  title={Slic-hf: Sequence likelihood calibration with human feedback},
  author={Zhao, Yao and Joshi, Rishabh and Liu, Tianqi and Khalman, Misha and Saleh, Mohammad and Liu, Peter J},
  journal={arXiv preprint arXiv:2305.10425},
  year={2023}
}

@article{xu2024contrastive,
  title={Contrastive preference optimization: Pushing the boundaries of llm performance in machine translation},
  author={Xu, Haoran and Sharaf, Amr and Chen, Yunmo and Tan, Weiting and Shen, Lingfeng and Van Durme, Benjamin and Murray, Kenton and Kim, Young Jin},
  journal={arXiv preprint arXiv:2401.08417},
  year={2024}
}

@inproceedings{huanglarger25,
  title={Larger or Smaller Reward Margins to Select Preferences for LLM Alignment?},
  author={Huang, Kexin and Wu, Junkang and Chen, Ziqian and Wang, Xue and Gao, Jinyang and Ding, Bolin and Wu, Jiancan and He, Xiangnan and Wang, Xiang},
  booktitle={Forty-second International Conference on Machine Learning},
  year={2025}
}

@inproceedings{muldrew2024active,
  title={Active Preference Learning for Large Language Models},
  author={Muldrew, William and Hayes, Peter and Zhang, Mingtian and Barber, David},
  booktitle={International Conference on Machine Learning},
  pages={36577--36590},
  year={2024},
  organization={PMLR}
}

@inproceedings{corrado2025automixalign,
  title={AutoMixAlign: Adaptive data mixing for multi-task preference optimization in LLMs},
  author={Corrado, Nicholas E and Katz-Samuels, Julian and Devraj, Adithya M and Yun, Hyokun and Zhang, Chao and Xu, Yi and Pan, Yi and Yin, Bing and Chilimbi, Trishul},
  booktitle={Proceedings of the 63rd Annual Meeting of the Association for Computational Linguistics (Volume 1: Long Papers)},
  pages={20234--20258},
  year={2025}
}

@misc{open-llm-leaderboard-v2,
  author = {Clémentine Fourrier and Nathan Habib and Alina Lozovskaya and Konrad Szafer and Thomas Wolf},
  title = {Open LLM Leaderboard v2},
  year = {2024},
  publisher = {Hugging Face},
  howpublished = "\url{https://huggingface.co/spaces/open-llm-leaderboard/open_llm_leaderboard}",
}

@misc{eval-harness,
  author       = {Gao, Leo and Tow, Jonathan and Abbasi, Baber and Biderman, Stella and Black, Sid and DiPofi, Anthony and Foster, Charles and Golding, Laurence and Hsu, Jeffrey and Le Noac'h, Alain and Li, Haonan and McDonell, Kyle and Muennighoff, Niklas and Ociepa, Chris and Phang, Jason and Reynolds, Laria and Schoelkopf, Hailey and Skowron, Aviya and Sutawika, Lintang and Tang, Eric and Thite, Anish and Wang, Ben and Wang, Kevin and Zou, Andy},
  title        = {The Language Model Evaluation Harness},
  month        = 07,
  year         = 2024,
  publisher    = {Zenodo},
  version      = {v0.4.3},
  doi          = {10.5281/zenodo.12608602},
  url          = {https://zenodo.org/records/12608602}
}

@article{sun2025robust,
  title={Robust Preference Optimization via Dynamic Target Margins},
  author={Sun, Jie and Wu, Junkang and Wu, Jiancan and Zhu, Zhibo and Lu, Xingyu and Zhou, Jun and Ma, Lintao and Wang, Xiang},
  journal={arXiv preprint arXiv:2506.03690},
  year={2025}
}

@misc{Tunstall_The_Alignment_Handbook,
  author = {Tunstall, Lewis and Beeching, Edward and Lambert, Nathan and Rajani, Nazneen and Huang, Shengyi and Rasul, Kashif and Bartolome, Alvaro and M. Patiño, Carlos and M. Rush, Alexander and Wolf, Thomas},
  license = {Apache-2.0},
  title = {{The Alignment Handbook}},
  howpublished = {\url{https://huggingface.co/alignment-handbook}},
  year={2024}
}

@article{amini2024direct,
  title={Direct Preference Optimization with an Offset},
  author={Amini, Afra and Vieira, Tim and Cotterell, Ryan},
  journal={Findings of the Association for Computational Linguistics: ACL 2024},
  pages={9954--9972},
  year={2024},
  publisher={Association for Computational Linguistics}
}

@inproceedings{hong2024orpo,
  title={ORPO: Monolithic Preference Optimization without Reference Model},
  author={Hong, Jiwoo and Lee, Noah and Thorne, James},
  booktitle={2024 Conference on Empirical Methods in Natural Language Processing},
  year={2024},
  organization={Association for Computational Linguistics}
}

@inproceedings{park2024disentangling,
  title={Disentangling Length from Quality in Direct Preference Optimization},
  author={Park, Ryan and Rafailov, Rafael and Ermon, Stefano and Finn, Chelsea},
  booktitle={Findings of the Association for Computational Linguistics ACL 2024},
  pages={4998--5017},
  year={2024}
}

@inproceedings{liu2025lipo,
  title={Lipo: Listwise preference optimization through learning-to-rank},
  author = {Liu, Tianqi and Qin, Zhen and Wu, Junru and Shen, Jiaming and Khalman, Misha and Joshi, Rishabh and Zhao, Yao and Saleh, Mohammad and Baumgartner, Simon and Liu, Jialu and Liu, Peter J. and Wang, Xuanhui},
  booktitle={Proceedings of the 2025 Conference of the Nations of the Americas Chapter of the Association for Computational Linguistics: Human Language Technologies (Volume 1: Long Papers)},
  pages={2404--2420},
  year={2025}
}

@inproceedings{liu2025understanding,
  title={Understanding reference policies in direct preference optimization},
  author={Liu, Yixin and Liu, Pengfei and Cohan, Arman},
  booktitle={Findings of the Association for Computational Linguistics: NAACL 2025},
  pages={8022--8037},
  year={2025}
}

@inproceedings{tanggeneralized,
  title={Generalized Preference Optimization: A Unified Approach to Offline Alignment},
  author={Tang, Yunhao and Guo, Zhaohan Daniel and Zheng, Zeyu and Calandriello, Daniele and Munos, Remi and Rowland, Mark and Richemond, Pierre Harvey and Valko, Michal and Pires, Bernardo Avila and Piot, Bilal},
  booktitle={Forty-first International Conference on Machine Learning},
  year={2024}
}

@inproceedings{xiao2025infopo,
  title={InfoPO: On mutual information maximization for large language model alignment},
  author={Xiao, Teng and Ge, Zhen and Sanghavi, Sujay and Wang, Tian and Katz-Samuels, Julian and Versage, Marc and Cui, Qingjun and Chilimbi, Trishul},
  booktitle={Proceedings of the 2025 Conference of the Nations of the Americas Chapter of the Association for Computational Linguistics: Human Language Technologies (Volume 1: Long Papers)},
  pages={11699--11711},
  year={2025}
}

\appendix

\section{Revisiting Baseline Methods}
\label{sec:methods}

\begin{table*}[t]
    \caption{Various preference optimization objectives and hyperparameter search range.}
    \vspace{-0.8em}
    \label{tab:baseline}
    \centering
    \resizebox{\textwidth}{!}{
    \small
    \renewcommand{\arraystretch}{1.2}
    \begin{tabular}{lll}
    \toprule 
    \textbf{Method} & \textbf{Objective} & \textbf{Hyperparameter} \\
    
    \midrule 
    \rowcolor{gray!30}\multicolumn{3}{c}{\emph{Common Offline Preference Optimization Methods}} \\
    \midrule
    
    DPO~\citep{rafailov2023direct} & $-\log \sigma \left( \beta \log \frac{\pi_\theta(y_w|x)}{\pi_{\text{ref}}(y_w|x)} - \beta \log \frac{\pi_\theta(y_l|x)}{\pi_{\text{ref}}(y_l|x)}\right)$ & $\beta \in [0.01, 0.05, 0.1]$ \\ \midrule
    \multirow{2}{*}{SLiC~\cite{zhao2023slic}} & \multirow{2}{*}{$\max (0,\delta -\log{\pi_\theta }(y_w|x)+\log{\pi_\theta }(y_l|x))-\lambda \log{\pi_\theta }(y_w|x)$} & $\lambda \in [0.1, 0.5, 1.0, 10.]$ \\
    & & $\beta \in [0.1, 0.5, 1.0, 2.0]$ \\ \midrule
    IPO~\citep{azar2024general} & $ \left( \log \frac{\pi_\theta(y_w|x)}{\pi_{\text{ref}}(y_w|x)} - \log \frac{\pi_\theta(y_l|x)}{\pi_{\text{ref}}(y_l|x)} - \frac{1}{2\tau} \right)^2$ & $\tau \in [0.01, 0.1, 0.5, 1.0]$ \\  \midrule 
    CPO~\citep{xu2024contrastive} &  $-\log \sigma  \left(\beta \log \pi_\theta(y_w|x) - \beta \log \pi_\theta(y_l|x) \right) - \lambda \log \pi_\theta (y_w|x)$ & $\alpha = 1.0, \,\, \beta \in [0.01, 0.05, 0.1]$ \\ \midrule
    \multirow{2}{*}{KTO~\citep{ethayarajh2024kto}} & $-\lambda_w \sigma \left( \beta \log \frac{\pi_\theta(y_w|x)}{\pi_{\text{ref}}(y_w|x)} - z_{\text{ref}} \right) +  \lambda_l \sigma \left( z_{\text{ref}} - \beta \log \frac{\pi_\theta(y_l|x)}{\pi_{\text{ref}}(y_l|x)} \right),\,$ & $\lambda_l = \lambda_w = 1.0$ \\  
    & $\text{where} \,\, z_{\text{ref}} = \mathbb{E}_{(x, y) \sim \mathcal{D}} \left[\beta \text{KL}\left( \pi_\theta(y|x) || \pi_{\text{ref}}(y|x) \right)  \right]$ & $\beta \in [0.01, 0.05, 0.1]$ \\ \midrule
    \multirow{2}{*}{SimPO~\citep{meng2024simpo}} & \multirow{2}{*}{$-\log \sigma  \left( \frac{\beta}{|y_w|} \log \pi_\theta(y_w|x) - \frac{\beta}{|y_l|} \log \pi_\theta(y_l|x) - \gamma \right)$} & $\beta \in [2.0, 4.0, 6.0, 8.0]$ \\
    & & $\gamma \in [0.3, 0.5, 1.0, 1.2, 1.4, 1.6]$ \\
    
    \midrule 
    \rowcolor{gray!30}\multicolumn{3}{c}{\emph{Solutions for Mitigating Hyperparameter Sensitivity}} \\
    \midrule
    
    \multirow{2}{*}{$\alpha$-DPO~\cite{wu2025alphadpo}} & {$-\log \sigma \left( u(x,y_w,y_l) - \operatorname{sg} \left[ \gamma + \alpha M^\ast(x, y_w, y_l) \right] \right)$} & $\beta\in [2.5,10.0], \,\, \gamma \in [0.1, 0.3, 0.5]$  \\
    & $\text{where} \,\, u(x,y_w,y_l)=\frac{\beta}{|y_w|} \log \pi_\theta(y_w|x) - \frac{\beta}{|y_l|} \log \pi_\theta(y_l|x)$ & $\alpha \in [1e-2, 5e-2, 0.1, 0.2]$ \\ \midrule
    \multirow{2}{*}{$\beta$-DPO~\citep{wu2024beta}} & {$-\log \sigma \left( \beta_i \log \frac{\pi_\theta(y_w|x)}{\pi_{\text{ref}}(y_w|x)} - \beta_i \log \frac{\pi_\theta(y_l|x)}{\pi_{\text{ref}}(y_l|x)}\right)$} & $\beta =0.1$, $\alpha \in [0.4, 0.6, 0.8, 1.0]$  \\
    & $\text{where} \,\, \beta_i=[1+\alpha(M_i-M_0)]\beta_0 $ & $\rho \in [0.0, 0.1, 0.2, 0.3, 0.4, 0.5]$ \\ \midrule
    \multirow{2}{*}{$\epsilon$-DPO~\cite{lee2025kl}} & {$-\log \sigma \left( \beta_i \log \frac{\pi_\theta(y_w|x)}{\pi_{\text{ref}}(y_w|x)} - \beta_i \log \frac{\pi_\theta(y_l|x)}{\pi_{\text{ref}}(y_l|x)}\right)$} & $\beta =0.01$ \\
    & $\text{where} \,\, \beta_i \leftarrow \{\beta_\epsilon^-,\beta_\epsilon^+,\beta\} $ & $\epsilon \in [0.005, 0.01, 0.02]$ \\ 
    
    \midrule 
    \rowcolor{gray!30}\multicolumn{3}{c}{\emph{Solutions for Eliminating Hyperparameter Sensitivity}} \\
    \midrule
    
    SimPER~\cite{xiao2025simper} & $-\mathrm{exp}\left(\frac{1}{|y_w|}\log \pi_\theta (y_w|x) \right) +\mathrm{exp}\left(\frac{1}{|y_l|}\log \pi_\theta (y_l|x) \right)$  & - \\ \midrule
    \multirow{2}{*}{\RMiPO} & $-\log \sigma  \left( \frac{\beta}{|y_w|} \log \pi_\theta(y_w|x) - \frac{\beta}{|y_l|} \log \pi_\theta(y_l|x) - \gamma(x,y_w,y_l) \right)$ & \multirow{2}{*}{$\beta = 2.0 \,\, \gamma_{\min}=0.3 \,\, \gamma_{\max}=1.6$} \\
    & $\text{where} \,\, \gamma(x, y_w, y_l) = \gamma_{\min} + (\gamma_{\max} - \gamma_{\min}) \cdot \exp(-\max(0, \Delta \mathrm{pmi})) $ &  \\
    \bottomrule
    \end{tabular}
    }
    \vspace{-.8em}
\end{table*}

This section revisits SOTA baseline methods for offline preference optimization by examining the underlying design motivations and the functional roles of hyperparameters in their loss formulations. Based on this analysis, we summarize and contrast the resulting optimization objectives in Table~\ref{tab:baseline}.

\begin{itemize}
    \item \textbf{DPO}~\citep{rafailov2023direct}: By introducing a difference term of log-likelihood ratios, DPO implicitly models the reward function, thereby avoiding reliance on an explicit reward model in RLHF. This method introduces a tunable hyperparameter \colorbox{gray!30}{$\beta$} to control the deviation of the policy model from the reference model, enabling a controllable trade-off between preference alignment and policy updates.

    \item \textbf{SLiC}~\cite{zhao2023slic}: The optimization objective of SLiC incorporates a supervised fine-tuning (SFT) term and is governed by two hyperparameters: the margin parameter \colorbox{gray!30}{$\delta$}, which sets the boundary of the ranking loss, and the weighting coefficient \colorbox{gray!30}{$\lambda$}, which balances the contribution of the SFT term within the overall objective.

    \item \textbf{IPO}~\citep{azar2024general}: IPO transforms preference learning into a regression problem with a squared loss by defining a surrogate reward function, thereby avoiding the instability associated with reinforcement learning training. This approach involves a hyperparameter \colorbox{gray!30}{$\beta$} that controls the magnitude of the reward signal.

    \item \textbf{CPO}~\citep{xu2024contrastive}: CPO directly uses the log-likelihood as the reward signal and jointly optimizes it with a SFT term. This method introduces two hyperparameters: \colorbox{gray!30}{$\beta$}, which scales the log-probability term, and \colorbox{gray!30}{$\lambda$}, which adjusts the weight of the SFT term in the joint objective.

    \item \textbf{KTO}~\citep{ethayarajh2024kto}: KTO learns from unpaired preference data, with its optimization objective governed by three hyperparameters: \colorbox{gray!30}{$\beta$}, which controls the degree of deviation of the policy model from the reference model, and \colorbox{gray!30}{$\lambda_w$} and \colorbox{gray!30}{$\lambda_l$}, which weight the preference terms corresponding to winning and losing responses, respectively.

    \item \textbf{SimPO}~\citep{meng2024simpo}: SimPO does not rely on a reference model and is trained by optimizing a length-regularized objective over response pairs, with its objective function governed by two hyperparameters: \colorbox{gray!30}{$\beta$}, which scales the log-probability term, and \colorbox{gray!30}{$\gamma$}, which adjusts the reward margin.
\end{itemize}

A review of the above paradigms reveals that hyperparameters are ubiquitous across different methods; as demonstrated in prior studies, their tuning is not only unavoidable but also often incurs substantial computational overhead. To address the issue of hyperparameter dependence, researchers have recently proposed several approaches.

\begin{itemize}
    \item \textbf{$\alpha$-DPO}~\cite{wu2025alphadpo}: $\alpha$-DPO aims to combine the advantages of DPO and SimPO to reduce sensitivity to hyperparameters \colorbox{gray!30}{$\beta$} and \colorbox{gray!30}{$\gamma$}. This approach introduces a smoothing parameter \colorbox{gray!30}{$\alpha \in [0,1]$} to interpolate between the two objectives: when $\alpha = 0$, the optimization objective reduces to SimPO; when $\alpha = 1$, it is equivalent to DPO. Central to SimPO is the length-normalized utility $u(x,y_w,y_l)=\frac{\beta}{|y_w|} \log \pi_\theta(y_w|x) - \frac{\beta}{|y_l|} \log \pi_\theta(y_l|x)$, which mitigates the bias introduced by response length. $\alpha$-DPO incorporates this utility into its unified objective. The method is based on the preference scoring function $M(x, y_w, y_l) = \beta \log \frac{\pi_\theta(y_w \mid x) \, \pi_{\mathrm{ref}}(y_l \mid x)}{\pi_{\mathrm{ref}}(y_w \mid x) \, \pi_\theta(y_l \mid x)}$, which measures the relative difference between the policy $\pi_\theta$ and the reference $\pi_{\mathrm{ref}}$ for a preference pair $(y_w, y_l)$. To enhance stability, $M(x, y_w, y_l)$ is Z-score normalized to obtain $M^\ast(x, y_w, y_l)$.

    \item \textbf{$\beta$-DPO}~\citep{wu2024beta}: $\beta$-DPO proposes dynamically adjusting the hyperparameter \colorbox{gray!30}{$\beta$} based on the quality of preference pairs to mitigate sensitivity and reduce tuning burden. Specifically, the quality of a preference pair $(y_w, y_l)$ is measured by $M_i = \beta_0 \log \frac{\pi_\theta(y_w \mid x)}{\pi_{\mathrm{ref}}(y_w \mid x)} - \beta_0 \log \frac{\pi_\theta(y_l \mid x)}{\pi_{\mathrm{ref}}(y_l \mid x)}$. A scaling factor \colorbox{gray!30}{$\alpha \in [0,1]$} is then introduced to modulate the influence of $M_i$ on the update of $\beta_i$. To further enhance robustness, the method incorporates a data filtering mechanism controlled by the hyperparameter \colorbox{gray!30}{$\rho$}, which suppresses the adverse effects of outlier preference pairs during $\beta$ selection.

    \item \textbf{$\epsilon$-DPO}~\cite{lee2025kl}: $\epsilon$-DPO examines the behavior of preference pairs under hyperparameter perturbations and proposes an adaptive mechanism for regulating \colorbox{gray!30}{$\beta$}. Specifically, a perturbation scale \colorbox{gray!30}{$\epsilon > 0$} is introduced, defining the bounds $\beta_{\epsilon}^{-} := \frac{\beta}{1+\epsilon}, \beta_{\epsilon}^{+} := \beta(1+\epsilon)$, which correspond to values slightly below and above $\beta$, respectively. It has been observed that the preference likelihood of the policy model exhibits local monotonicity around $\beta$. Based on this property, $\epsilon$-DPO dynamically adjusts $\beta$ for each sample within the neighborhood determined by $\epsilon$, thereby enhancing the robustness and adaptability of the optimization.

    \item \textbf{SimPER}~\cite{xiao2025simper}: SimPER demonstrates that optimizing only the inverse perplexity is sufficient for strong performance, where inverse perplexity is defined as the reciprocal of the exponentiated average log-likelihood of the chosen versus rejected responses in the preference dataset. Based on this, the method constructs a \colorbox{gray!30}{hyperparameter-free} objective, offering advantages in computational and memory efficiency.

\end{itemize}

The above solutions for hyperparameter tuning have all demonstrated significant effectiveness. However, an analysis of their optimization objectives indicates that, except for SimPER, existing methods primarily alleviate the tuning burden by reducing the number of tunable hyperparameters or narrowing the search space. In practical deployment, we observe that although SimPER has a simple structure and strong performance, its training stability is relatively limited. \emph{Therefore, an ideal method should completely eliminate hyperparameter dependence while maintaining good stability.}

\section{Full Derivations}
\label{sec:proofs}

\noindent \textbf{Derivation of Insight I}. To analyze the impact of hyperparameters on offline preference optimization methods such as DPO and SimPO, we unify them under the following objective function framework:
$\mathcal{L}(\theta) = -\mathbb{E}_{(x, y_w, y_l) \sim \mathcal{D}} \left[ \log \sigma \left( \beta \Delta \log - \gamma \right) \right]$,
where $\Delta \log := \log \pi_\theta(y_w \mid x) - \log \pi_\theta(y_l \mid x)$ denotes the log-likelihood difference of the policy model $\pi_\theta$ on a preference triplet $(x, y_w, y_l)$. Taking the gradient with respect to $\theta$ and applying the chain rule yields:
\begin{equation}
\begin{aligned}
&\nabla_\theta \mathcal{L} 
 = -\mathbb{E} \Bigg[ \frac{1}{\sigma(\beta \Delta \log - \gamma)} 
\cdot \sigma(\beta \Delta \log - \gamma) \\
&\cdot \big(1 - \sigma(\beta \Delta \log - \gamma)\big)
\cdot \nabla_\theta (\beta \Delta \log - \gamma)
\Bigg].
\end{aligned}
\end{equation}
Since $\gamma$ is a constant independent of $\theta$, we have $\nabla_\theta \gamma = 0$. After simplification, the gradient can be rewritten as:
\begin{equation}
\begin{aligned}
\nabla_\theta \mathcal{L} = -\beta \cdot \mathbb{E} \left[ \underbrace{\big( 1 - \sigma(\beta \Delta \log - \gamma) \big)}_{\text{Weight } w_i} \nabla_\theta \Delta \log \right]
\end{aligned}
\end{equation}
Here, $w_i = 1 - \sigma(\beta \Delta \log - \gamma)$ can be interpreted as the effective weight of the $i$-th sample in the gradient update. Stationary points satisfy $\nabla_\theta \mathcal{L} = 0$. Because $\beta > 0$, this factor can be canceled in the first-order necessary condition, yielding a characterization of the stationary set $\mathcal{S}$:
\begin{equation}
\begin{aligned}
&\mathbb{E} \left[ w_i \nabla_\theta \Delta \log \right] \\
&= \mathbb{E} \left[ \big( 1 - \sigma(\beta \Delta \log - \gamma) \big) \nabla_\theta \Delta \log \right] = 0.
\end{aligned}
\end{equation}
To clarify the distinct roles of $\beta$ and $\gamma$ in convergence behavior, we examine the structure of the weight term $w_i$. Specifically, the symmetry center of the sigmoid function is determined by $\beta \Delta \log = \gamma$, at which point $w_i = 0.5$. This indicates that $\gamma$ directly governs the relative importance of different samples in the expected gradient, thereby systematically shifting the stationary set $\mathcal{S}$. In contrast, $\beta$ acts as a global scaling factor on the gradient magnitude, controlling the steepness of the sigmoid function, i.e., the sharpness of the decision boundary. Although $\beta$ substantially influences optimization dynamics, such as convergence speed and robustness to noise, it does not alter the sample-importance structure dominated by $\gamma$ in the first-order stationary condition. \emph{Thus, $\gamma$ plays a more fundamental role than $\beta$ in shaping the stationary points of the optimization.}

\noindent \textbf{Derivation of Insight II}. As discussed previously, the hyperparameter $\gamma$ plays a central role in preference optimization objectives by regulating the importance of samples. However, if this modulation signal itself is influenced by bias in the prior distribution of responses, it may cause the model to overfit high-frequency yet weakly semantically related responses, thereby degrading alignment quality and generalization ability. Therefore, in the ideal case, there is an urgent need for a modulation signal that reflects purely semantic relevance and is insensitive to response popularity.

Current mainstream methods directly use the log-likelihood difference $\Delta \log$ of the policy model as the optimization signal. However, according to Bayes’ theorem, the conditional probability $\pi_{\theta}(y|x)$ can be decomposed as $\pi_{\theta}(y|x) = \frac{\pi_{\theta}(x|y) \cdot \pi_{\theta}(y)}{\pi_{\theta}(x)}$. Taking the logarithm yields:
\begin{equation}
\log \pi_{\theta}(y|x) = \log \pi_{\theta}(x|y) + \log \pi_{\theta}(y) - \log \pi_{\theta}(x).
\end{equation}
Correspondingly, the log-likelihood difference $\Delta \log$ for a preference pair $(x, y_w, y_l)$ can be decomposed as:
\begin{equation}
\begin{aligned}
\Delta \log ={}& \underbrace{\left( \log \pi_{\theta}(x|y_w) - \log \pi_{\theta}(x|y_l) \right)}_{\text{Semantic Relevance Gap}} \\
&+ \underbrace{\left( \log \pi_{\theta}(y_w) - \log \pi_{\theta}(y_l) \right)}_{\text{Prior Popularity Bias}}.
\end{aligned}
\end{equation}
Here, the term $\log \pi_{\theta}(y_l)$ characterizes the model’s prior preference for response $y$, which is typically highly correlated with its frequency in the training data. The associated risk is that high-frequency responses with generic content are often excessively labeled as “preferred,” even when their semantic association with the input is weak. If the optimization objective directly relies on $\Delta \log$, the model will tend to further amplify the likelihood of such popular responses, leading to degradation in semantic–input alignment and triggering downstream issues such as reduced generation diversity and weakened factual consistency~\cite{li2024contrastive}.

\begin{algorithm*}[t]
\caption{\RMiPO: Response-level mutual information for preference optimization}
\label{alg:rmipo}
\begin{algorithmic}[1]
\renewcommand{\baselinestretch}{1.1}\selectfont
\REQUIRE Preference dataset $\mathbb{D}$, Batch size $b$, Hyperparameters $\beta, \gamma_{\min}, \gamma_{\max}$, Learning rate $\eta$
\STATE Initialize policy parameters $\pi_{\theta}$ 
\WHILE{not converged}
    \STATE Sample a batch of preference pairs $\{(x^{(i)}, y^{(i)}_w, y^{(i)}_l)\}_{i=1}^b$ from $\mathbb{D}$
    
    \STATE \texttt{// Step 1: Compute the relevant probabilities based on the policy model.}
    
    \STATE Compute the policy probabilities for the preferred response $y_w$ and the less-preferred response $y_l$: $\pi_\theta(y_w|x), \pi_\theta(y_w), \pi_\theta(y_l|x), \pi_\theta(y_l)$.

    \STATE Compute the modulation signal $\Delta \mathrm{pmi}$ using the formulation in Equation~\ref{eq:pmi}.
    
    \STATE \texttt{// Step 2: Compute the length-normalized reward difference}
    \STATE $\Delta \mathbb{R} = \frac{\beta}{|y_w|} \log \pi_\theta(y_w | x) - \frac{\beta}{|y_l|} \log \pi_\theta(y_l | x)$

    \STATE \texttt{// Step 3: Compute the reward margin $\gamma$ corresponding to each triplet.}
    \STATE $\gamma(x, y_w, y_l) = \gamma_{\min} + (\gamma_{\max} - \gamma_{\min}) \cdot \exp(-\max(0, \Delta \mathrm{pmi}))$

    \STATE \texttt{// Step 4: Optimization Update}
    \STATE Compute the loss $\mathcal{L}_{\text{\RMiPO}}$ using the formulation in Equation~\ref{eq:loss}
    \STATE Compute the gradient and update the policy $\theta_t \leftarrow \theta_{t-1} - \eta \nabla_\theta \mathcal{L}_{\text{\RMiPO}}(\theta_{t-1})$
\ENDWHILE
\RETURN Optimized policy $\pi_\theta$
\end{algorithmic}
\end{algorithm*}

To eliminate this bias, this paper introduces pointwise mutual information (PMI) as a decoupling tool. PMI is defined as $\text{PMI}(y|x) = \log \frac{\pi_{\theta}(y|x)}{\pi_{\theta}(y)}$. Clearly, PMI measures the additional information gain brought by response $y$ given input $x$, and it naturally removes the prior probability $\pi_{\theta}(y)$ of the response itself. Based on this, we define the adaptive modulation signal for a preference pair as the PMI difference: $\Delta \text{pmi} = \text{PMI}(y_w|x) - \text{PMI}(y_l|x)$. Substituting the expression of PMI and simplifying yields:
\begin{equation}
\begin{aligned}
\Delta \text{pmi} ={}& \left( \log \pi_{\theta}(x|y_w) - \log \pi_{\theta}(x) \right) \\
&- \left( \log \pi_{\theta}(x|y_l) - \log \pi_{\theta}(x) \right) \\
={}& \log \pi_{\theta}(x|y_w) - \log \pi_{\theta}(x|y_l).
\end{aligned}
\end{equation}
It follows that $\Delta \text{pmi}$  depends solely on the semantic relevance gap, effectively eliminating the prior bias from $\pi_{\theta}(y)$. \emph{An adaptive $\gamma$ modulation based on $\Delta \text{pmi}$ can improve generalization performance and generation quality in real-world scenarios.}

\section{Implementation Details}
\label{sec:details}

\noindent \textbf{Dataset}: In this work, the training of the policy model primarily relies on two preference datasets: UltraFeedback-Binarized~\cite{cui2023ultrafeedback} and Anthropic-HH~\cite{bai2022training}.

\begin{itemize}
    \item \textbf{UltraFeedback-Binarized}: The original UltraFeedback dataset contains 64,000 prompts, each with four responses from various open- and closed-source language models. All responses are rated by GPT-4 based on overall quality (overall\_score). In the UltraFeedback-Binarized version, a preference pair is formed by the highest-scoring response (the “preferred” response) and one randomly selected response from the remaining three (the “less-preferred” response). Policy models trained on this dataset are evaluated on instruction-following benchmarks and downstream tasks.
    \item \textbf{Anthropic-HH}: Anthropic-HH is a human-labeled dialogue preference dataset comprising two subsets, focusing on “helpfulness” and “harmlessness.” The dataset originates from manually constructed and red-teamed dialogue samples, designed to study how crowd-sourced annotators perform red-team attacks and which strategies are effective. Policy models trained on Anthropic-HH are primarily used for safety alignment evaluation.
\end{itemize}

\noindent \textbf{Training}: \RMiPO~is designed for simplicity and strong performance, with the full procedure provided in Algorithm~\ref{alg:rmipo}. Its core lies in leveraging response-level PMI for hyperparameter adaptation. Specifically, $\pi_{\theta}(y)$ denotes the prior popularity of response $y$, defined as the joint probability of the sequence: $\pi_{\theta}(y) = \prod_{t=1}^{|y|} \pi_{\theta}(y_t \mid y_{<t})$. Unlike $\pi_{\theta}(y \mid x)$, this formulation removes dependence on the prompt $x$, enabling explicit decoupling between response priors and semantic relevance.

For probability estimation, $\pi_{\theta}(y)$ is computed exactly over the full sequence without truncation or approximation, ensuring consistency with information-theoretic definitions. For datasets where prompts and responses are concatenated, we apply a lightweight string-matching heuristic to extract $y$. Although this may introduce minor boundary inaccuracies, it preserves intended decoupling. Consequently, $\Delta \text{pmi}$ serves as an intrinsic measure of semantic separability, mitigating bias from non-semantic factors such as high-frequency tokens.

All models are trained under a unified setup with $\beta = 2.0$, $\gamma_{\min} = 0.3$, and $\gamma_{\max} = 1.6$, following the search space of SimPO. During SFT, we use a learning rate of $2 \times 10^{-5}$; across both SFT and preference optimization, we adopt a batch size of 128, a maximum sequence length of 2048, and train for one epoch using Adam with cosine annealing. All experiments are conducted on four 80GB A800 GPUs. Baselines follow the same search protocol as SimPO: method-specific hyperparameters are detailed in Table~\ref{tab:baseline}, and learning rates are tuned within $[3 \times 10^{-7}, 1 \times 10^{-6}]$. In contrast, \RMiPO~(and SimPER) require tuning only the learning rate, without introducing additional hyperparameters.

\noindent \textbf{Evaluation}: Evaluation on AlpacaEval 2 adopts the same hyperparameter settings as SimPO: decoding temperatures are set to 0.7 and 0.5 for Mistral-Base and Mistral-Instruct, respectively, while both Llama-3 variants use a temperature of 0.9. Evaluation on MT-Bench follows the official decoding configuration, which specifies task-specific sampling temperatures. To systematically examine the impact of preference optimization methods on downstream task performance, we evaluate models trained with each method on the standard tasks included in the Hugging Face Open LLM Leaderboard, strictly following established evaluation protocols. For safety evaluation, we assess models trained on the Anthropic HH dataset using its test set as the benchmark, measuring the probability that the model generates preferred responses over less-preferred ones.

\section{Extended Experiments}
\label{sec:exps}

\setlength{\tabcolsep}{4pt}
\begin{table}[t]
\centering·
\small 
\caption{Without length-normalization on AlpacaEval2.}
\label{tab:LN-AE2}
\resizebox{\columnwidth}{!}{
\begin{tabular}{l cc cc}
\toprule
\multirow{1}{*}{\textbf{Method}} 
& \multicolumn{2}{c}{\textbf{Mistral-7B-Instruct}}  
& \multicolumn{2}{c}{\textbf{Llama3-8B-Instruct}} 
\\ 
\cmidrule(lr){2-3}\cmidrule(lr){4-5}
~w/o~LN
& {LC (\%)} 
& {WR (\%)}  
& {LC (\%)}  
& {WR (\%)}  \\
\midrule
DPO  
& 20.3 & 18.0 & 40.2 & 38.1
\\
SimPO 
& 30.5 & 38.2 & 42.4 & 40.4
\\
$\alpha$-DPO 
& 32.0 & 38.4 & \textbf{44.4} & 42.6
\\
SimPER 
& 28.5 & 34.6 & 41.7 & 40.2
\\
\RMiPO
& \textbf{33.7} & \textbf{38.8} & \textbf{44.4} & \textbf{43.1} 
\\
\bottomrule
\end{tabular}
}
\vspace{-.8em}
\end{table}

\noindent \textbf{Length-normalized Performance Analysis}.
In this work, length-normalized training is treated as an auxiliary technique for improving training stability rather than a core contribution. Although SimPO shows that length normalization improves model performance, particularly in terms of length-controlled win rate, we further examine the inherent stability and effectiveness of \RMiPO~without relying on this technique. To this end, we conduct experiments on \RMiPO~and several baselines with Llama-3-8B-Instruct and Mistral-7B-Instruct. As shown in Table~\ref{tab:LN-AE2}, the results indicate that even without length normalization, \RMiPO~consistently yields improvements that are substantially stronger than those of the baselines.

\begin{table*}[!h]
    \centering
    \ttfamily
    \small
        \begin{tabular}{lp{13.5cm}}
        \toprule[1pt]
        \bf{Prompt} & To ensure a more comprehensive approach, the \#Given Prompt\# should now include a diverse range of cultural backgrounds and political perspectives. This will require thorough analysis and understanding of various societal norms and beliefs, emphasizing the criticality of cross-cultural communication and sensitivity towards different ideologies.\\
        \midrule
        \bf{\RMiPO} & 
        The revised prompt may read:\newline
        
        "Create a dialogue within a multicultural community consisting of individuals from at least three distinct cultural backgrounds and varying political affiliations, discussing the challenges and opportunities presented by globalization in today's interconnected world, while addressing concerns related to economic inequality, cultural assimilation, communal identity, and environmental sustainability." \\
    
        \midrule
        \bf{SimPER} & 
        It is crucial to approach this prompt with an open mind and a willingness to learn about perspectives that may differ from one's own. The end goal is to create a comprehensive and well-rounded essay that addresses the given prompt from multiple cultural and political perspectives.\\

        \midrule
        \bf{SimPO} & 
        The resulting written piece should strive to accurately represent and explore multiple perspectives while remaining coherent and engaging to a broad audience.\\
        
        \midrule
        \bf{\textsc{gpt-4o-mini}} & 
        To ensure a more comprehensive approach, the \#Given Prompt\# should actively incorporate a diverse range of cultural backgrounds and political perspectives. This will involve: \newline
        
        1. \textbf{Researching Cultural Norms}: Understanding the foundational beliefs, values, and customs that shape various societies. This includes exploring different family structures, social hierarchies, and community practices that influence individual and collective behaviors.\newline
        
        2. \textbf{Exploring Political Perspectives}: Analyzing different political ideologies such as liberalism, conservatism, socialism, and others prevalent in specific regions. Acknowledging how these ideologies impact governance, social policies, and public opinion in various countries.\newline
        
        3. \textbf{Cross-Cultural Communication}: Emphasizing the importance of understanding and respecting cultural differences in communication styles. This includes recognizing the significance of non-verbal cues, context, and the role of language in shaping interactions across cultures.\newline
        
        4. \textbf{Sensitivity Towards Ideologies}: Being aware of and respecting differing beliefs and values, particularly when they conflict with one’s own. This requires an open-minded approach, active listening, and a willingness to understand perspectives that may be contrary to personal views.\newline
        
        5. \textbf{Diverse Representation}: Ensuring that voices from various backgrounds are included in discussions, thereby enhancing the richness of the conversation. This can involve engaging with community leaders, cultural representatives, and activists from different political spectra.\newline
        
        6. \textbf{Empathy and Understanding}: Cultivating empathy by placing oneself in the shoes of individuals from various backgrounds. This helps in appreciating the complexities of their experiences and the factors that shape their viewpoints.\newline
        
        7. \textbf{Critical Reflection}: Encouraging self-reflection on one’s biases and preconceived notions. This is crucial for fostering an inclusive dialogue that values all perspectives and contributions.\newline
        
        By integrating these elements, the \#Given Prompt\# will promote a fuller understanding of the multifaceted nature of human experience and the importance of fostering inclusive environments where diverse perspectives can coexist and enrich dialogue.\\
        \bottomrule[1pt]
    \end{tabular}
    \caption{Comparative analysis of response quality on the UltraFeedback test set. For identical prompts, the Mistral-7B-Base optimized with \RMiPO~produces significantly higher-quality and more helpful outputs than SOTA methods.}
    \vspace{-.8em}
    \label{tab:case}
\end{table*}

\noindent \textbf{Generated Response Case Analysis}.
To more intuitively demonstrate the performance advantages of \RMiPO~in practical generation tasks, we select a complex instruction involving “multicultural background and political viewpoint balance” from the UltraFeedback test set as an illustrative case. Table~\ref{tab:case} presents a comparative illustration of outputs from the Mistral-7B-Base model optimized with \RMiPO, SimPER, and SimPO, with results from \textsc{gpt-4o-mini} provided for qualitative reference.

The analysis indicates that \RMiPO~exhibits significant advantages in terms of instruction-following and operational depth. While the prompt asks for a "comprehensive approach" to include diverse backgrounds, \RMiPO~is the only model that successfully translates this abstract meta-instruction into a concrete, executable prompt. It proposes a structured dialogue involving at least three distinct cultures and specifically identifies intersectional themes such as globalization, economic inequality, and environmental sustainability. This demonstrates a shift from simple "passive inclusion" to "active, high-fidelity cultural synthesis." In contrast, SimPO and SimPER remain at a superficial or meta-discursive level: SimPO provides a high-level summary of the desired outcome but fails to generate a functional revised prompt, offering only a general goal of being "coherent and engaging." SimPER focuses on the attitude and mindset of the writer, providing ethical guidelines rather than a structured response to the task itself. This case clearly substantiates the quantitative results reported in Table~\ref{tab:ae2_mtb}: \RMiPO~not only outperforms state-of-the-art baselines in instruction-following win rates, but also enhances the model’s reasoning capability and structured expression ability when addressing high-difficulty, multi-dimensional tasks.

\begin{figure}
    \centering
    \includegraphics[width=\columnwidth]{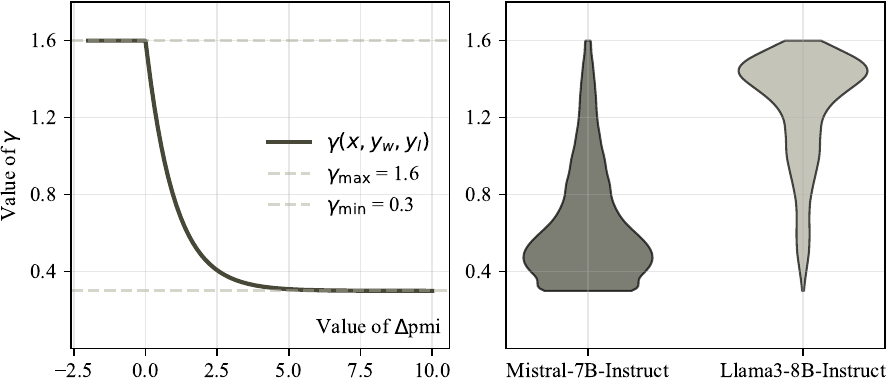}
    \caption{Adaptive $\gamma$ Analysis. \textbf{Left}: The adaptive mechanism $\gamma(x, y_w, y_l)$ follows an exponential decay based on $\Delta \mathrm{pmi}$, dynamically scaling the learning signal for each instance. \textbf{Right}: Empirical distribution of $\gamma$ values on Mistral-7B and Llama3-8B. The density peaks (approximately 0.4 and 1.4, respectively) are approximately consistent with the optimal fixed values reported by SimPO, supporting the effectiveness of \RMiPO's automated, instance-level parameter optimization.}
    \vspace{-0.5cm}
    \label{fig:gamma}
\end{figure}

\noindent \textbf{Adjusted $\gamma$ Distribution Analysis}.
To investigate the adaptive alignment mechanism of \RMiPO, this paper systematically analyzes the modulation coefficient $\gamma$ from two perspectives—mechanism design and empirical distribution (see Figure~\ref{fig:gamma}).

First, at the level of mechanism design, \RMiPO~introduces a dynamic decay strategy based on pointwise mutual information differences between samples. As shown in the left subfigure of Figure~\ref{fig:gamma}, $\gamma$ is defined as follows: $\gamma(x, y_w, y_l) = \gamma_{\min} + (\gamma_{\max} - \gamma_{\min}) \cdot \exp(-\max(0, \Delta \mathrm{pmi}))$. This mechanism dynamically adjusts the value of $\gamma$ according to the pointwise mutual information difference ($\Delta \mathrm{pmi}$) between positive and negative samples. When the discriminability of a sample pair is low, $\gamma$ remains at a high level ($\gamma_{\max} = 1.6$) to strengthen the learning signal; as the difference increases, $\gamma$ smoothly decays toward the lower-bound value ($\gamma_{\min} = 0.3$) following an exponential decay pattern. This design enables fine-grained weight allocation across examples of varying difficulty. \emph{We emphasize that this functional form serves as a practical instantiation of our principle, rather than the only valid implementation. While other monotonic decay functions could be used, we adopt the exponential form for superior smoothness and empirical stability.}

Second, the empirical distribution of $\gamma$ observed during training validates the effectiveness and robustness of this adaptive mechanism. As shown in the right subfigure of Figure~\ref{fig:gamma}, upon completion of training, we compute the corresponding $\gamma$ value for each sample in the dataset based on the final policy model. The results show that for Mistral-7B-Instruct, the density of $\gamma$ is mainly concentrated around 0.4, whereas for Llama3-8B-Instruct, the primary distribution range is close to 1.4. Notably, these automatically emerging distribution centers are closely aligned with the optimal fixed parameter values obtained by SimPO on the same models through extensive grid search.

These results demonstrate that \RMiPO~effectively replaces manual hyperparameter tuning with a principled adaptive parameterization mechanism. By transforming hyperparameters that traditionally require manual tuning into internal adaptive variables, \RMiPO~not only eliminates costly hyperparameter search, but also achieves more precise alignment through instance-level dynamic weight adjustment. This instance-level adaptation is a key reason why \RMiPO~consistently maintains strong performance across diverse model architectures.

\noindent \textbf{Space and Time Complexity Analysis}.
To evaluate the practicality of \RMiPO~and baseline methods, we analyze space consumption (GPU memory) and time efficiency (computational throughput) under identical hardware settings, parallelization strategies, and model architectures.

\textit{Space complexity}. GPU memory during training is dominated by model parameters, optimizer states, gradients, and activations. DPO-based methods (e.g., DPO, $\beta$-DPO, $\epsilon$-DPO) require an additional frozen reference model $\pi_{\text{ref}}$ to compute the KL constraint. Although it does not require optimizer states, its parameters still occupy substantial memory, leading to a peak utilization of 82\%. In contrast, SimPO, SimPER, and \RMiPO~eliminate the reference model entirely. As a result, \RMiPO~reduces peak memory usage to 62\% (an approximately 20\% absolute reduction), enabling larger batch sizes or longer sequences and mitigating memory and communication bottlenecks in distributed training.

\textit{Time complexity}. Per-step runtime is determined by the number of forward and backward passes. DPO-style methods incur additional latency due to an extra forward pass through $\pi_{\text{ref}}$, although this overhead is limited because no gradients are computed. SimPO and SimPER require only a single forward and backward pass, achieving the highest efficiency. In contrast, $\alpha$-DPO introduces complex dynamic offsets and nonlinear gradient operations, resulting in the highest per-step latency. \RMiPO~has a runtime of $O(T) + \Delta$, where $\Delta$ arises from computing the PMI-based margin term $\gamma(x, y_w, y_l)$ with lightweight logit-level operations. Despite this small overhead, \RMiPO~converges faster and achieves better alignment.

Compared with DPO, which incurs an additional 30\%–50\% memory overhead, \RMiPO~maintains a lower memory footprint via its reference-free design while introducing only minimal computational overhead, yielding a favorable trade-off between efficiency, scalability, and alignment performance.

\textbf{Mechanism analysis}.
This section provides additional theoretical clarification and empirical validation for the exponential scheduling mechanism and constraint bounds adopted in \RMiPO. We aim to further substantiate that these design choices are grounded in principled considerations rather than heuristic tuning.
Preference optimization is inherently conducted in the log-probability space. Therefore, an exponential mapping naturally aligns with the log-ratio structure of the objective function. This alignment ensures that the adaptive scaling remains geometrically consistent with the additive formulation in the log domain. More specifically, the exponential transformation preserves the relative ordering of preference signals while enabling controlled amplification of marginal differences. Compared with alternative mappings (e.g., linear or cosine), it provides a more balanced modulation behavior: in high-confidence regions, it avoids under-modulation and preserves discriminative gradients; in low-confidence regions, it suppresses excessive scaling and prevents noisy updates. As a result, the exponential schedule facilitates a smoother and more stable gradient flow throughout training.
The parameters $\gamma_{\min}$ and $\gamma_{\max}$ are introduced as stability constraints rather than task-specific hyperparameters. Their primary role is to restrict the adaptive scaling within a safe region. These bounds serve two key purposes: preventing gradient explosion during early training stages or under noisy PMI estimation, and avoiding the dominance of marginal scaling over the main optimization objective. From a regularization perspective, the constraint interval enforces robustness across different random seeds and model architectures, ensuring consistent convergence behavior.

\emph{Ablation on Scheduling Strategies}.
We compare four scheduling strategies: \textit{Exponential}, \textit{Linear}, \textit{Cosine}, and a baseline without explicit scheduling (\textit{w/o}). Experiments are conducted on the AlpacaEval 2 benchmark under the length-control (LC) setting. The results are summarized in Table~\ref{tab:schedule}. The exponential mapping consistently outperforms alternative strategies across both models. This suggests that the nonlinear curvature of the exponential mapping is more effective at capturing subtle preference distinctions in complex semantic comparisons, thereby supporting our geometric-consistency hypothesis.

\begin{table}[h]
\centering
\small
\begin{tabular}{lcccc}
\toprule
Model & Exponential & Linear & Cosine & w/o \\
\midrule
Mistral-7B-Instruct & 38.1 & 36.9 & 35.5 & 37.2 \\
Llama-8B-Instruct   & 49.1 & 47.9 & 46.5 & 47.2 \\
\bottomrule
\end{tabular}
\caption{Win rates (\%) under different scheduling strategies on AlpacaEval 2 (LC setting).}
\label{tab:schedule}
\end{table}

\emph{Sensitivity Analysis on Constraint Bounds}.
We further evaluate the sensitivity of RMiPO to different choices of $(\gamma_{\min}, \gamma_{\max})$. The results are summarized in Table~\ref{tab:gamma}. The performance remains stable across a wide range of boundary settings, with only minor fluctuations. This robustness suggests that the bounds primarily function as safety constraints rather than sensitive tuning parameters.

\begin{table}[h]
\centering
\small
\begin{tabular}{lcccc}
\toprule
Model & $[0, 2]$ & $[0.3, 1.6]$ & $[0.5, 1.4]$ & $[1.0, 1.2]$ \\
\midrule
Mistral-7B-Instruct & 38.5 & 38.1 & 37.4 & 36.9 \\
Llama-8B-Instruct   & 50.4 & 49.1 & 48.3 & 47.5 \\
\bottomrule
\end{tabular}
\caption{Sensitivity analysis of constraint bounds.}
\label{tab:gamma}
\end{table}

Overall, both theoretical analysis and empirical results demonstrate that:
(i) the exponential scheduling mechanism is well-aligned with the log-space structure of preference optimization; and
(ii) constraint bounds act as regularization safeguards, ensuring stable and robust training dynamics.
These findings confirm that the proposed design is principled and generalizable, rather than relying on heuristic parameter choices.

\end{document}